\documentclass[a4paper,fleqn]{cas-dc}

\usepackage[numbers]{natbib}
\usepackage{xcolor}
\usepackage{subcaption}

\begin{document}
\let\WriteBookmarks\relax
\def\floatpagepagefraction{1}
\def\textpagefraction{.001}

\shorttitle{Boosting-inspired online learning with transfer for railway maintenance}

\shortauthors{Risca et~al.}

\title [mode = title]{Boosting-inspired online learning with transfer for railway maintenance}


\tnotetext[1]{Work funded by Portuguese Foundation for Science and Technology under project doi.org/10.54499/UIDP/00760/2020 and Ph.D. scholarship PRT/BD/154713/2023. It also received EU funds, through Portuguese Republic’s Recovery and Resilience Plan, within project PRODUTECH R3.}

%

\author[1]{Diogo Risca}[orcid=0009-0000-1495-0662]

\author[1]{Afonso Louren\c{c}o}[orcid=0000-0002-3465-3419]

\cormark[1]


\ead{fonso@isep.ipp.pt}



\affiliation[1]{organization={GECAD, ISEP, Polytechnic of Porto},
    addressline={Rua Dr. António Bernardino de Almeida}, 
    city={Porto},
    postcode={4249-015}, 
    country={Portugal}}
   


\author[1]{Goreti Marreiros}[orcid=0000-0003-4417-8401]

\cortext[cor1]{Corresponding author}


\begin{abstract}
The integration of advanced sensor technologies with deep learning algorithms has revolutionized fault diagnosis in railway systems, particularly at the wheel-track interface. Although numerous models have been proposed to detect irregularities such as wheel out-of-roundness, they often fall short in real-world applications due to the dynamic and nonstationary nature of railway operations. This paper introduces BOLT-RM (\textbf{B}oosting-inspired \textbf{O}nline \textbf{L}earning with \textbf{T}ransfer for \textbf{R}ailway \textbf{M}aintenance), a model designed to address these challenges using continual learning for predictive maintenance. By allowing the model to continuously learn and adapt as new data become available, BOLT-RM overcomes the issue of catastrophic forgetting that often plagues traditional models. It retains past knowledge while improving predictive accuracy with each new learning episode, using a boosting-like knowledge sharing mechanism to adapt to evolving operational conditions such as changes in speed, load, and track irregularities. The methodology is validated through comprehensive multi-domain simulations of train-track dynamic interactions, which capture realistic railway operating conditions. The proposed BOLT-RM model demonstrates significant improvements in identifying wheel anomalies, establishing a reliable sequence for maintenance interventions.
\end{abstract}

\begin{keywords}
Railway systems \sep Fault diagnosis \sep Continual learning \sep Predictive maintenance \sep Deep learning \sep Experience replay 
\end{keywords}

\maketitle

\section{Introduction}
\label{chap:Chapter1}

Ensuring the safety of passengers and freight is paramount in the railway industry, requiring a focus on reliability and maintainability \cite{Hidirov2019}. Although trains are designed to operate for decades, maintaining their safety and quality at an optimal level while minimizing costs is a challenge due to high traffic, massive axle loads, and constantly changing conditions. The wheel-rail interface, in particular, represents a significant portion of maintenance costs \cite{Frohling2010}. Components such as wheels, rails, and sleepers are continuously exposed to environmental and operational conditions, leading to issues such as corrosion, cracks, and other damage \cite{vitorGoncalves2023, Guedes2023, Mosleh2021}. Furthermore, wheel-rail surfaces endure high contact stresses and sliding in rolling contact, with an increasing risk function commonly observed in failure rates \cite{Chong2010, Mohammadi2023}.

\textbf{Data-driven maintenance.} Industries have traditionally relied on preventive maintenance, which schedules periodic interventions to avoid failures, and corrective maintenance, which acts only after a failure occurs. However, the increasing probability of failure, i.e. hazard function, in wheel-rail components highlights the need for a more data-driven predictive maintenance (PdM) approach. Based on on-board and side condition monitoring, PdM uses data analytics and machine learning (ML) to predict failures and schedule maintenance before they occur. This reduces downtime and costs, and extends the remaining useful life of components \cite{Alemi2017, moleda2023, Pech2021}.

\begin{figure}[ht]
    \centering
    \vspace{7pt}
    \includegraphics[width=0.4\textwidth]{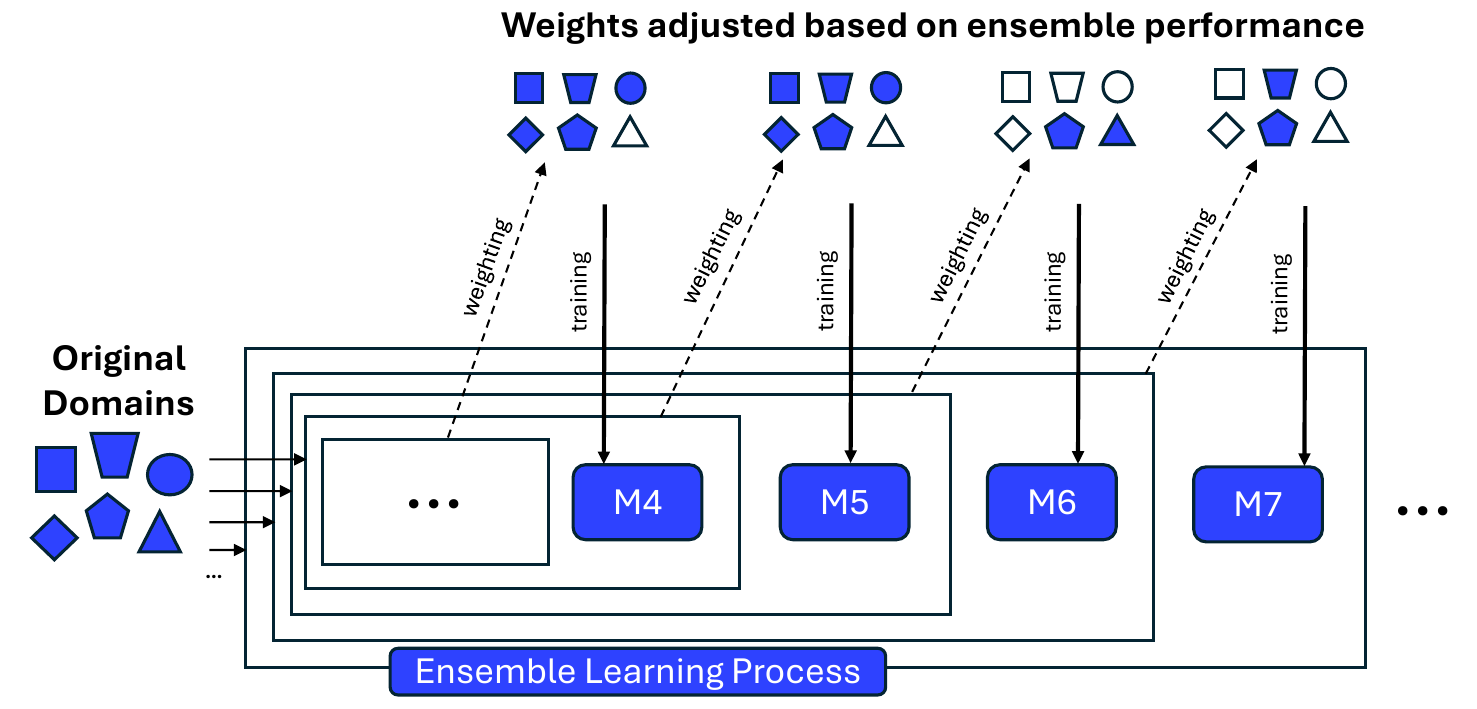}
    \caption{Boosting-like knowledge sharing}
    \label{fig:boosting}
\end{figure}

\textbf{Time series.} Across most railway PdM use cases, the most common data structure are time series, which is extremely challenging for standard modeling techniques. The changing mean, variance, trends, seasonality, and residual noise pose inherent non-stationary complexities, complicating models that assume identical distributions. Furthermore, autocorrelation introduces dependencies between observations, complicating models that assume independent data points. Missing or irregularly spaced data further exacerbate these issues, requiring robust preprocessing methods that respect temporal dependencies, with anomaly detection today increasing the likelihood of similar faults in the near future due to wear and tear. These challenges highlight the need for advanced methods to accurately extract and interpret time series information \cite{Torres2021, Seymour1997}.

\textbf{Feature extraction.} A common solution are advanced feature extraction techniques. For example, \cite{Mohammadi2023} propose an unsupervised learning method using wayside condition monitoring, using techniques like autoregressive (AR) models, PCA, CWT, and ARX to enhance sensitivity and accuracy under varying conditions. Similarly, \cite{Lourenco2023} developed an adaptive time series representation method that integrates Hidden Markov Models (HMM) for localized feature extraction, effectively isolating defective wheels and assessing defect severity. Wavelet-based approaches for early flat wheel detection are explored in \cite{Mosleh2023}, focusing on timely fault detection to ensure operational safety. Additionally, \cite{lourencco2023online} employ a query-by-content approach using Dynamic Time Warping (DTW) to rank wheel passages with anomalies.

\begin{figure}[ht]
    \centering
    \begin{subfigure}[b]{0.4\textwidth}
        \centering
        \includegraphics[width=\textwidth]{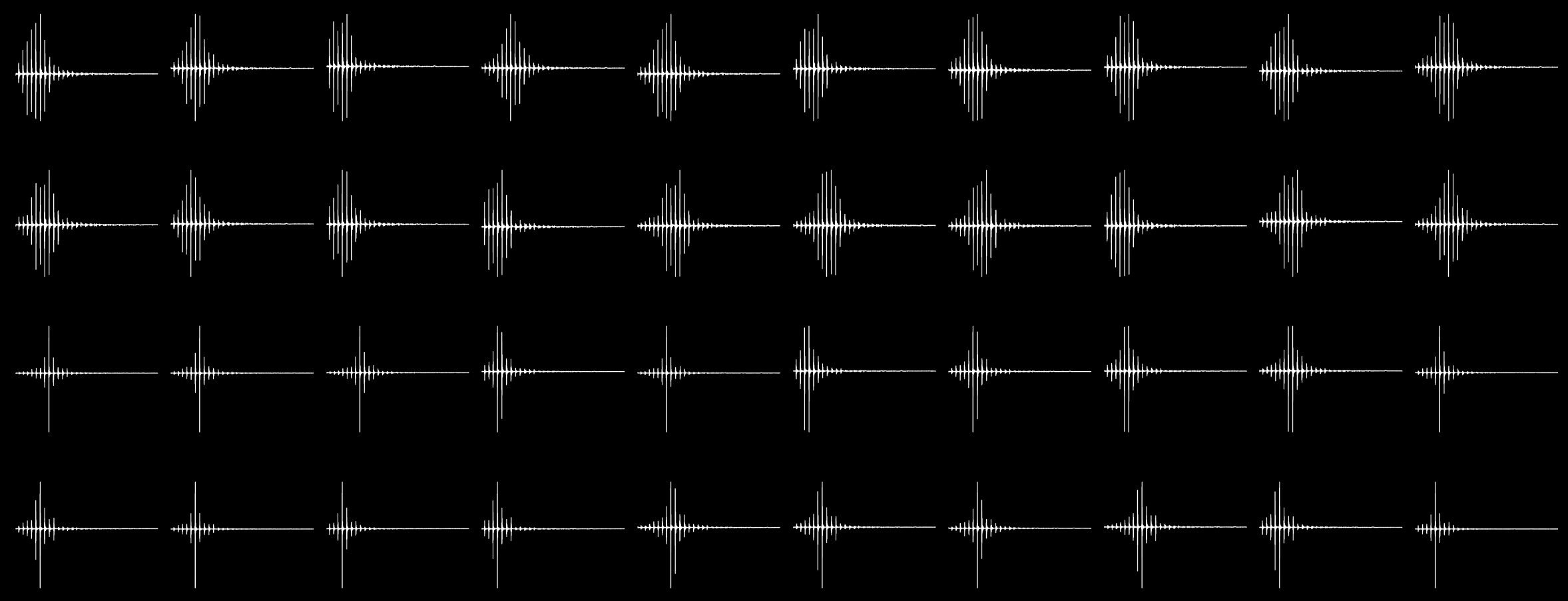}
        \subcaption{1D signal}
    \end{subfigure}
    \hfill
    \begin{subfigure}[b]{0.4\textwidth}
        \centering
        \includegraphics[width=\textwidth]{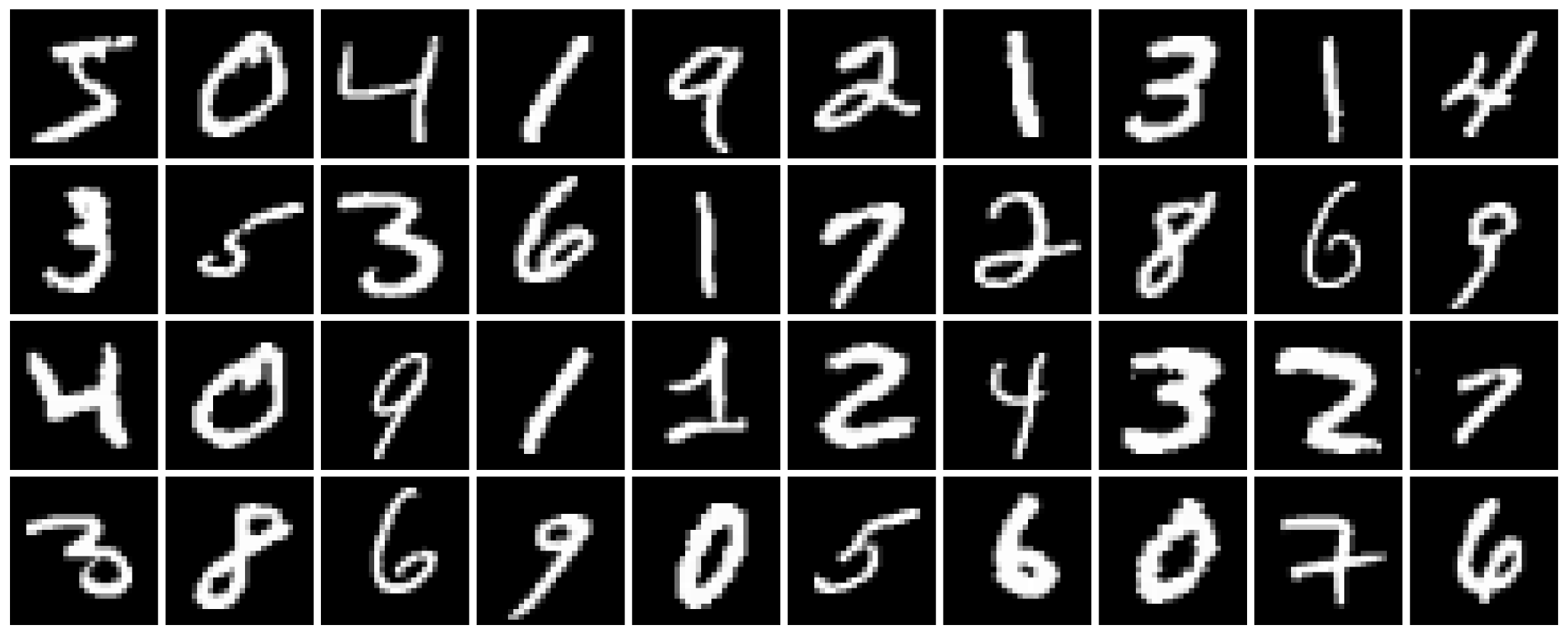}
        \subcaption{MNIST}
    \end{subfigure}
    \hfill
    \begin{subfigure}[b]{0.4\textwidth}
        \centering
        \includegraphics[width=\textwidth]{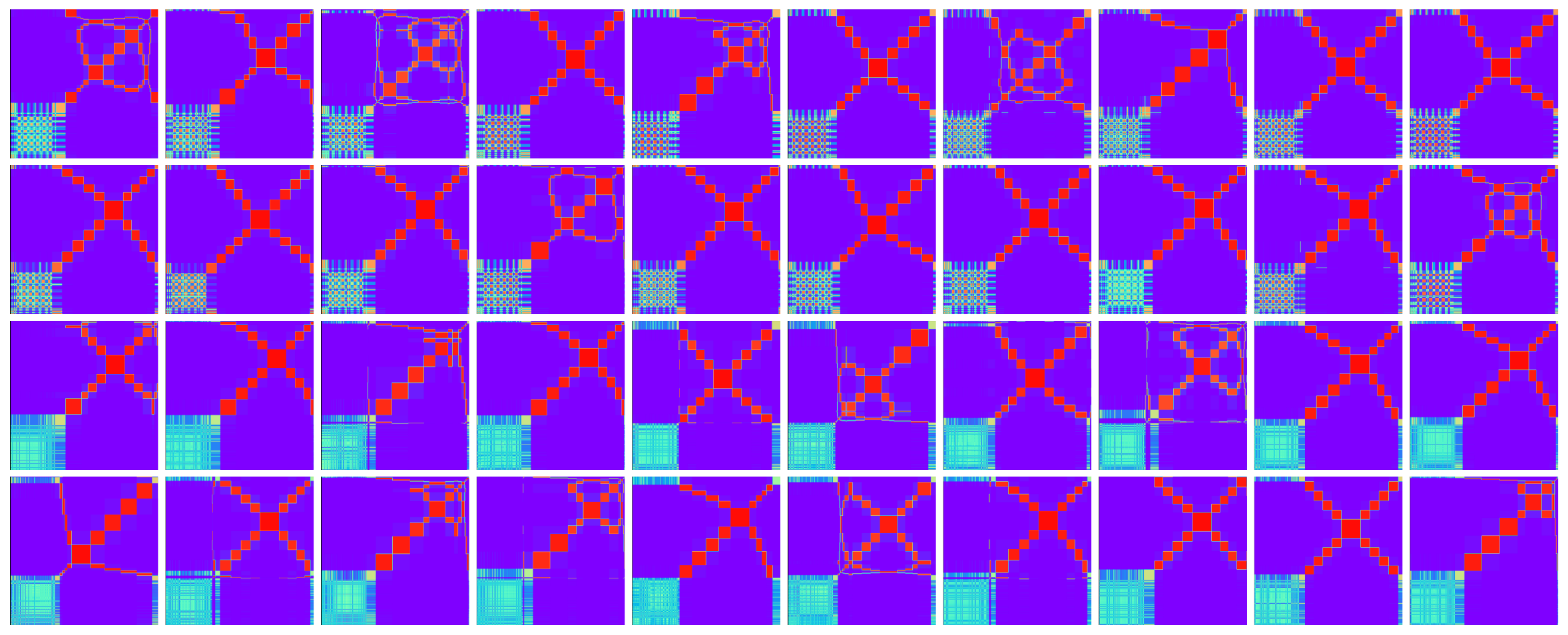}
        \subcaption{2D signal}
    \end{subfigure}
    
    \caption{Vision data}
    \label{fig:mtfimages}
\end{figure}

\textbf{Visual patterns.} However, such techniques often have difficulty in truly uncovering anomalies in signals, which are easily apparent to experienced observers through visual inspection. Comparison in spikes or irregularities across multiple examples of signals easily reveals variations indicative of wheel damage to a trained eye, as shown in Figure~\ref{fig:mtfimages}. Similarly to how MNIST visually treats handwritten digits \cite{lecun2010mnist}, one can visually learn and recognize patterns and shapes within a grid-like topology of time series data \cite{LeCun2015, Schmidhuber2015}. For instance, one can apply 1D convolution on the time axis, with CNNs learning temporal features and capturing dependencies at multiple time scales through convolutional filters, and pooling layers for dimensionality reduction \cite{Patil2021}. Alternatively, one can apply a pre-processing operation to encode temporal dependencies into a 2D matrix suitable for CNN-based analysis \cite{Hassan2019}.

\begin{figure}[ht]
    \centering
    \includegraphics[width=0.47\textwidth]{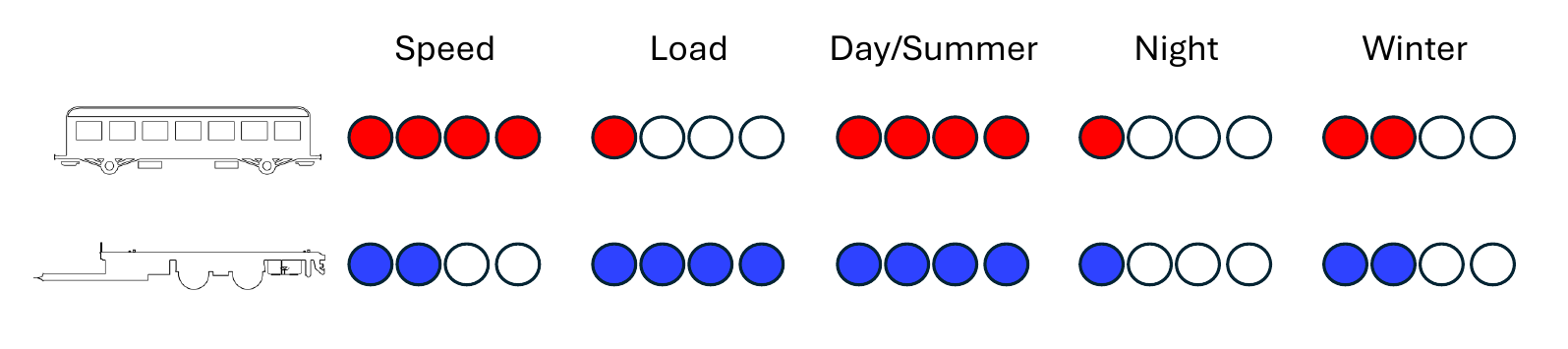}
    \caption{Domain-specific EOVs}
    \label{fig:env_trains}
\end{figure}

\textbf{Batch learning.} Beyond dealing with the challenge of capturing a time series structure, the dynamic nature of railway operations also requires going beyond traditional predictive models \cite{lourencco2024time}. Rail conditions are never static, as their state is in constant flux due to weather, wear, tear, and traffic variation. A fault diagnosis system must adapt to new wheel wear patterns caused by changing track conditions, weather, or load variations, as illustrated in Figure \ref{fig:env_trains}. For instance, high-speed trains with heavier loads experience greater impacts from wheel-rail irregularities, while lighter, slower trains exhibit different tolerance thresholds. As these operational and environmental domains evolve, the system must detect anomalies not present in the previous data. Thus, requiring models that capture the distinct dynamics of each train's operational profile.

\textbf{Continual learning.} To adress this, algorithms must enable learning from continuous data streams, adapting to any newly available information (plasticity) while keeping the knowledge that has already been acquired (stability) \cite{Hurtado2023}. For this purpose, a wide array of methods have been developed, such as regularization-based strategies, i.e., EWC, synaptic intelligence, and learning without forgetting \cite{Maschler2021}. However, despite the extensive literature, its application to railway systems remains limited. An exception is \cite{Liu2024} that proposed WindTrans, a transformer-based model for short-term forecasting of wind speeds in high-speed railway systems, where environmental factors significantly influence maintenance needs. It combines a graph encoder for spatial correlations and a temporal decoder for long-sequence modeling, along with an experience replay scheme to adapt to evolving wind patterns.

\begin{figure}[ht]
    \centering
    \includegraphics[width=0.4\textwidth]{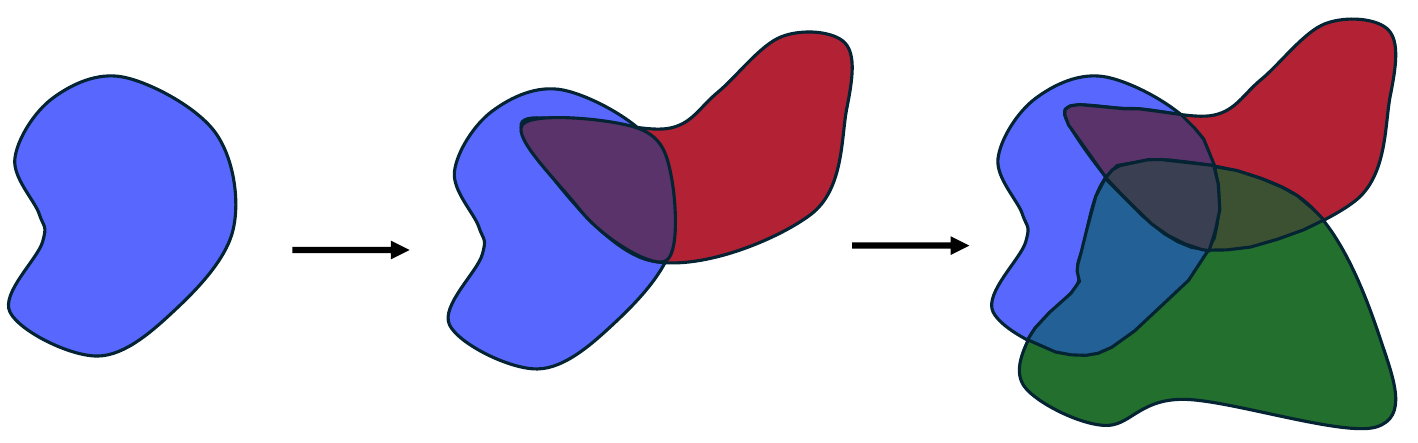}
    \caption{NP-hard}
    \label{fig:modelSharing}
\end{figure}

\textbf{Shared capacity.} Moreover, such efforts in CL mostly focus on invariant shared solutions \cite{bouvier:tel-03663398}, relying on replay and regularization for managing the stability-plasticity dilemma, but inevitably facing stability issues due to inter-domain interference, which worsens as the number of domains increases and their similarity decreases. Learning such invariant representations risks compromising adaptability, akin to addressing incremental domains within a constrained parameter space, which has been proven to be NP-hard as the feasible parameter space becomes increasingly narrow and irregular with additional domains \cite{Knoblauch2020}, as illustrated in Figure \ref{fig:modelSharing}.

\begin{figure}[ht]
    \centering
    \includegraphics[width=0.23\textwidth]{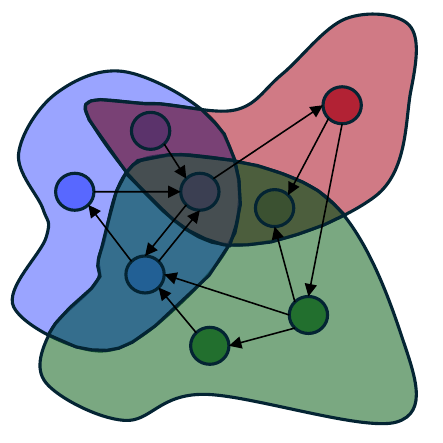}
    \caption{Modularity}
    \label{fig:modularity}
\end{figure}

\textbf{Modular capacity.} Indeed, model sharing across domains exists on a spectrum, with full sharing and no sharing as two extremes, and compositionality as a middle ground. Grouping domains allows for controllability in knowledge sharing between these groups, playing a crucial role in enabling the model to generalize well in a wide range of operational scenarios. As represented in Figure \ref{fig:modularity}, each domain shares information with others, allowing new domains to benefit not only from internal knowledge but also from the insights gained by overlapping or adjacent domains. This process facilitates smoother transitions and faster learning when encountering new operational conditions. Related domains can use their similarity for positive transfer, where learning one domain enhances performance in another or simplifies its (re)learning. Such transfer can be observed forward, with a old domain aiding current domains, or backwards, with current domains benefiting previously learned ones \cite{lopezpaz2017}.

\textbf{Ensemble.} A simple yet effective solution is to rely on deep ensembles, that initialize and train multiple neural networks independently \cite{lakshminarayanan2017simple,fort2019deep}. This not only improves robustness, but also boosts the overall performance thanks to a higher diversity, heterogeneity, and de-correlated predictions \cite{havasi2020training}. However, while these are well known for supervised learning \cite{fort2019deep}, their functionality in continual learning scenarios has only recently been fully studied. In practice, most current applications focus on learning each single task with a sub-network \cite{rusu2016progressive,aljundi2017expert}, which can be seen as a mere instantiation of dynamic architectures, that expand the network structure when a new domain comes. For instance, \cite{rusu2016progressive} produces an additional network to perform on a new domain, which inherits the knowledge of old domains by connecting to previously learned networks, while \cite{yoon2017lifelong} expands the network architecture by adding a predetermined number of parameters and learning with sparse regularization for size optimization. However, such methods in this family inevitably face the limitation of storage, training and inference costs, that increase significantly as the number of models grows. To mitigate this, some recent works have observed that efficient ensembles of multiple continual learning models can bring huge benefits \cite{wang2022coscl,doan2023continual,caccia2022anytime}, while leveraging on advances of neural network subspaces \cite{wortsman2021learning} and mode connectivity \cite{garipov2018loss}. Not only modular adaptation of individual models leads to an attenuation of forgetting and a boost in the overall performance \cite{caccia2022anytime}, but also can help reduce extra parameter costs for task-specific sub-networks \cite{wen2020batchensemble} and save computational cost \cite{doan2023continual}.

\textbf{Domain relatedness.} Across such ensemble approaches, the key challenge lies in reformulating the forgetting problem into a task interference problem and solve it using model selection to discover cooperative domains. For this purpose, one can implicitly differentiate helpful and harmful knowledge based on structural allocation \cite{mallya2018packnet,serra2018overcoming,abati2020conditional}, i.e. assigning a disjoint subset of parameters to each task, thus, not suffering from updating old knowledge with a new one, e.g. using an architecture consisting of a global feature extractor and multiple head classifiers corresponding to domains. These disjoint sets of old tasks can be fixed \cite{mallya2018packnet} or rarely \cite{serra2018overcoming} updated while training a new task, while relying on the train-prune-retrain paradigm \cite{mallya2018packnet}, learning a mask \cite{mallya2018piggyback}, or attention through gradient descent \cite{serra2018overcoming}. However, these methods do not discover explicit relationships between tasks. To mitigate this issue, \cite{rypesc2024divide} rely on a mixture-of-experts approach, where each expert represents each class with a Gaussian distribution, and the optimal expert is selected to be fine-tuned, based on the similarity of those distributions, while \cite{abati2020conditional} rely on selecting an internal network structure with a channel-gating module. Alternatively, one can go even further, and introduce sensitivity measures to the loss of the current domain from the associated domains to find cooperative relations \cite{jin2022helpful}, and even emulate the boosting process for selecting domains to train with for each round \cite{Ramesh2022}.

\textbf{This work.} Building on these advancements, the study introduces BOLT-RM, a model designed for fault diagnosis in the wheel-rail interface. The main contributions are summarized as follows: 
\begin{itemize}
    \item Section~\ref{chap:Chapter2} details the fault diagnosis methodology used in this study, with raw time series data being transformed into Markov transition field (MTF) images, and a novel algorithm inspired by the hypothesis that humans organize learned knowledge by clustering similar domains, enabling efficient learning within clusters and rapid adaptation to new domains using existing knowledge or forming new clusters  \cite{Hsu2018, Bouton2004}. These design choices allow the model to adapt to various operational and environmental scenarios, applying knowledge from previously learned domains adapting across multiple domains.
    \item Section~\ref{chap:Chapter3} outlines a numerical modeling and simulation framework for the creation of a continual learning benchmark with different wheel-rail domains. It not only includes dynamic train-track interactions, but also different train types, speeds, loads, and track conditions.
    \item Section~\ref{chap:Chapter4} presents and analyzes the experimental results, showing BOLT-RM achieves an average domain accuracy of 93\%, compared to 54\% for the isolated model. Forward transfer reached a value of 0.73, while preserving a backward transfer score of nearly zero (\(3.47 \times 10^{-4}\)), confirming its ability for knowledge sharing while retaining previously acquired knowledge without performance degradation.
\end{itemize}

\section{BOLT-RM}
\label{chap:Chapter2}

\begin{figure}[ht]
    \centering
    \includegraphics[width=0.4\textwidth]{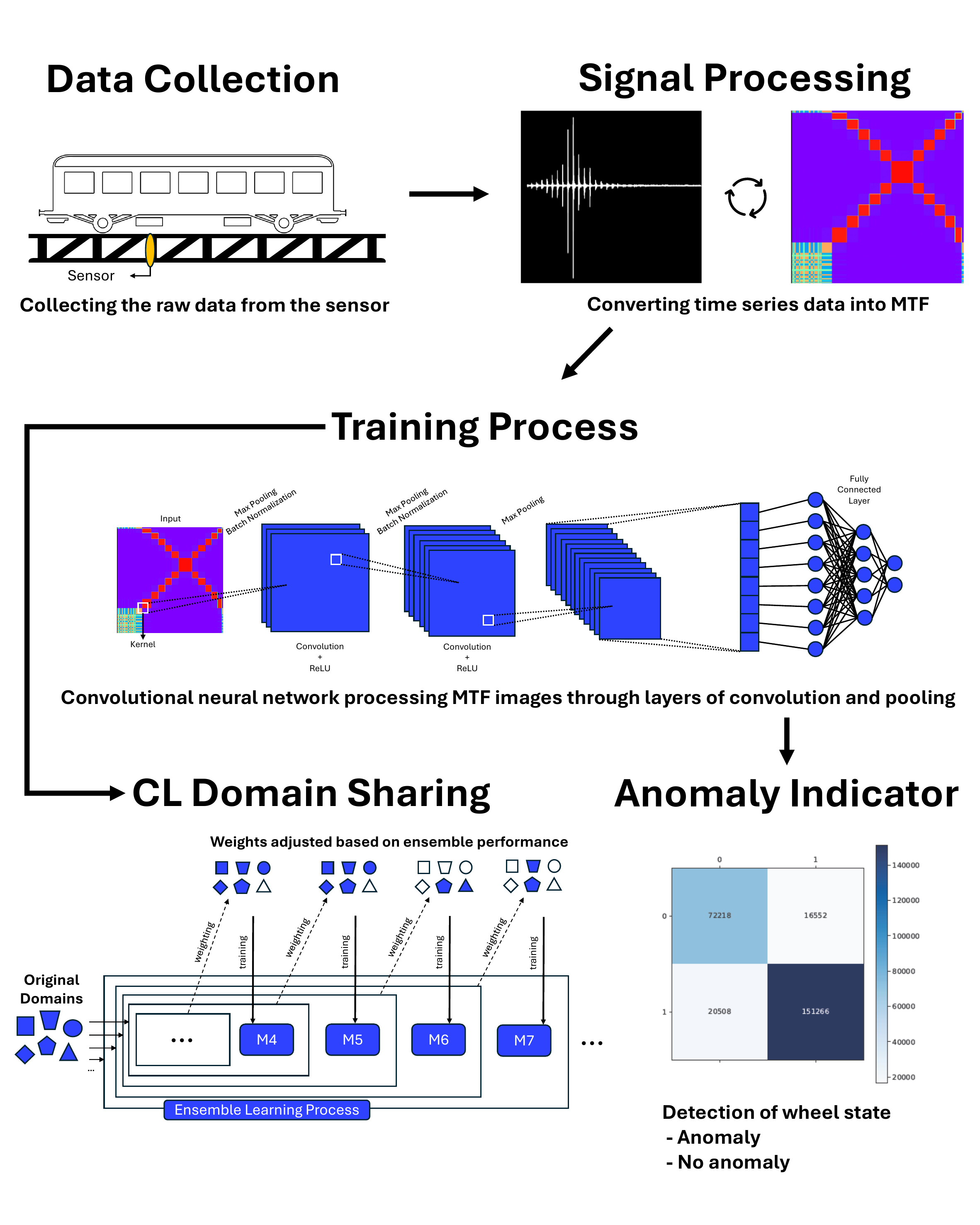}
    \caption{Railway wheel fault diagnosis}
    \label{fig:methodology_flowchart}
\end{figure}

This section outlines the frameworks and methodologies used for fault diagnosis on railway wheels, providing an overview of the methodology employed for damage detection. As illustrated in Figure \ref{fig:methodology_flowchart}, the process involves four key steps: data collection, signal processing, convolutional neural network (CNN) training, and anomaly detection. BOLT-RM is a specially designed ensemble learning paradigm for continual learning (CL) domains in which its capacity grows by adding smaller models for new domains. Hence, this framework is particularly well suited for the application, as it allows for adaptation to new data and changes patterns in wheel damage at any time.

\subsection{Vision-based signal processing}

The methodology involves placing sensors on the rail track to measure signals each time a train passes. These signals are traditionally analyzed as time series, and the raw data used are in that format. However, instead of directly using the image of the signal plot, the Markov transition field (MTF) spectrogram is adopted to encode the dynamics of the signal into a structured image. MTF captures the probability of state transitions over time, preserving both the temporal relationships and the underlying structure of the data. This transformation has proven to benefit various fault diagnosis applications \cite{Yan2022, Chen_2024, Lei2022, Wang2022}, whose findings indicate that MTF effectively maintains temporal relationships and offers a comprehensive depiction of state changes within the data, making it highly suitable for identifying subtle discrepancies required in fault detection. These principles are equally applicable to this research, where preserving temporal characteristics is essential.

\textbf{Markov transition field.} The MTF is derived from the transition matrix of quantized time series data. Consider a time series \(X = \{x_1, x_2, \ldots, x_T\}\) quantized into \(Q\) discrete states, where each value \(x_i\) is assigned to a corresponding state \(q_j\). From this quantization, a matrix \(W\) of size \(Q \times Q\) is constructed, where the element \(w_{ij}\) represents the probability that a value in state \(j\) is followed by a value in state \(i\). This probability satisfies the condition \(\sum_{j=1}^{Q} w_{ij} = 1\), ensuring that the matrix \(W\) represents a valid Markov transition matrix. However, while the matrix \(W\) encodes the state transition probabilities, it fails to incorporate sufficient information about the temporal structure and distribution of the original time series \(X\). This leads to a loss of important information, making \(W\) inadequate for capturing the full dynamics of the data. To address this limitation, the Markov transition field matrix \(M\) is introduced as an enhancement. The elements of \(M\) are constructed to include temporal information. For two states \(q_i\) and \(q_j\), the element \(M_{ij}\) represents the transition probability from state \(q_i\) to \(q_j\), conditioned on specific points in the time series. This is expressed as: $M_{ij} = w_{ij} \mid x_t \in q_i, x_{t+k} \in q_j$, where \(k\) represents the time interval between the two states. This allows \(M\) to incorporate both the state transition probabilities and their temporal dependencies. The full matrix \(M\) is constructed such that its elements reflect the probabilities of transitions between all pairs of states, accounting for various time intervals \(k\). Diagonal elements of \(M\), denoted as \(M_{ii}\), represent the special case where \(k = 0\), indicating the probability of remaining in the same state \(q_i\) at a specific time \(t\). By incorporating this temporal information, the Markov transition field matrix \(M\) effectively captures both the structural and dynamic characteristics of the original time series, providing a richer representation for subsequent analysis \cite{Yan2022, Han2021}.

\subsection{Model architecture}

The model architecture is operationalized to enable growth capacity in a sequence of domains, while sharing knowledge from previous domains. Assume that a sequence of domains $D_1, \ldots, D_n$ is presented to the system, each sharing the same input $X$, but different outputs $Y_1, \ldots, Y_n$. In each episode $k$, the model is tasked with training in the current domain $D_k$ and a selected subset of previous domains to facilitate knowledge sharing. For example, during episode $k = 2$, the training involves a feature generator $h$ and domain-specific classifiers, leading to the formation of the models $g_1$ $ \circ $ $h : X \rightarrow Y_1$ and $g_2$ $ \circ $ $h : X \rightarrow Y_2$. The model then classifies the inputs from both domains, producing a probability vector \( p_{g_i \circ h}(y | x), \forall y \in Y_i \) based on the respective domain. In episode \(k\), such set of domains can be defined as \(\overline{D}_k = \{D_{w_k^1}, \ldots, D_{w_k^b}\}\), where \(b \leq k\) serves as a hyper parameter, and \(\omega_k^i \in \{1, \ldots, k\}\). Training in \(\overline{D}_k\) involves the use in a feature generator $h_k$ and domain-specific classifiers \( g_{(k, \omega_k^i)} \) for each chosen domain. These models collectively form the current model, with the ability to predict data from $D_i$ for \(i \leq k\) being derived from averaging class probabilities output by all models that were applied to that domain, as indicated by:

\begin{equation}
    p_{k,i}(y|x) \propto \sum_{l=1}^{k} 1_{\{P_i \in \overline{P}_l\}} g_{l,i} \circ h_l(x)
    \label{dataprediction}
\end{equation}

\textbf{Domain-specific ensembling.} Such process is akin to using an ensemble of smaller CNNs. Unlike traditional models that try to handle all domains with a single larger model, this model divides domains into several smaller CNNs, with each one being trained in a specific domain or in a group of specific domains. This division helps manage the complexity of learning with multiple domains by dividing the workload, and ensuring that models do not interfere with each other's learning. With parameter isolation for each CNN, the model preserves the knowledge of each domain, which is essential to maintain good performance across all domains.

\textbf{Shared feature learning.} Subsequently, this ensemble of domain-specific CNNs relies on a shared feature generator, that pulls out meaningful features from the input data, acting not only as a filter, but also enhancing essential details of the input. For example, in analyzing the MTF images, the feature generator may effectively extract meaningful details from inputs derived from both strain and accelerometer signals. Accelerometer-based features, in particular, may be more sensitive to damage and less influenced by non-linear operational and environmental factors \cite{Lourenco2023}, making them highly valuable for anomaly detection. Conversely, strain-based features are excellent at preserving the fundamental shape of the feature, allowing to be better induce environmental and operational variations, such as the train type and wheel configuration. By using features from both types of signals, the model ensures a comprehensive analysis of potential faults. Moreover, this centralized feature extraction process is extremely cost-effective for continual learning with unbounded data streams, while at the same time allowing each classifier to specialize in its designated domain \cite{Tripuraneni2020}.

\subsection{Training process}

\begin{figure}[ht]
    \centering
    \includegraphics[width=0.47\textwidth]{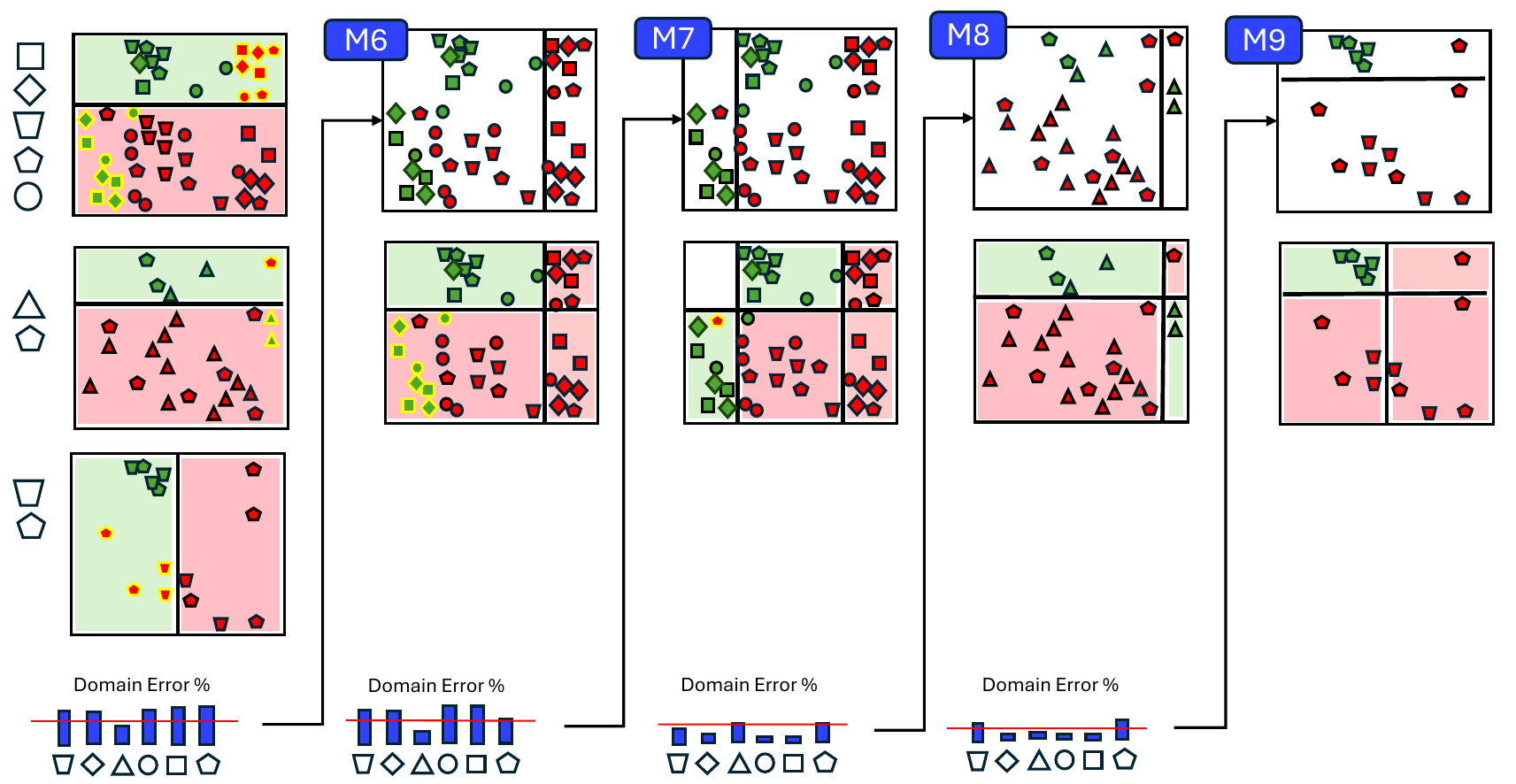}
    \caption{Domain selection and error-driven retraining process}
    \label{fig:training_process}
\end{figure}

The training procedure is inspired by boosting techniques, where multiple weak learners are combined to create a stronger learner. In traditional boosting, the training weights for each instance in the next episode are adjusted on the basis of the performance weaknesses of each individual model. However, in this approach, the weights for the next training episode are based on the performance of the entire ensemble up to that point, rather than on individual CNNs. This difference allows the model to adapt its learning more effectively across multiple domains considering the collective knowledge of the ensemble. Assuming that $\overline{w}_{k,i} \in \mathbb{R}^n$ is a normalized vector of domain-specific weights, after episode \(k\):
\begin{equation}
    \overline{w}_{k,i} \propto \exp\left(-1/m \sum\nolimits_{(x,y) \in S_i} \log p_{k,i}(y | x)\right)
\end{equation}
for each domain \(D_i\) with \(i \leq k\); for \(i > k, \overline{w}_{k,i} = 0\). Subsequently, in the following episode, domains \(\overline{D}_{k+1}\) are drawn from a multinomial distribution with weights \(\overline{w}_{k}\). With this, it makes it possible to put lower weight on domains with a lower empirical risk with for the next boosting episode. Thus, ensuring the system progressively concentrates on harder-to-classify domains, similarly to how AdaBoost reduces the training error by progressively focusing on difficult samples \cite{Schapire2013,Ramesh2022}. Figure \ref{fig:training_process} illustrates this process, where at each step, domains are evaluated based on their error percentages, with domains exceeding the error threshold prioritized for retraining in the subsequent episode. In BOLT-RM, however, the model incrementally learns a maximum of five domains, prioritizing those with the highest error percentages. This ensures continuous improvement while maintaining a comprehensive understanding of both newly introduced and previously learned domains.

\textbf{Experience replay.} Naturally, this training process requires revisiting a small fraction of data from previous domains. For this purpose, the architecture integrates a restricted data replay mechanism, with 4 key components that could be potentially tuned: (1) rehearsal representation, with one storing raw input data or latent features from hidden layers on a memory buffer \citep{hayes2021remind}, (2) label strategy, with memory buffer can holding a predicted label, such as logits \cite{buzzega2020dark}, instead of the true label, which is not always be feasible in real-time applications, (3) rehearsal policy, with identification of stored samples inside each domain for rehearsal using a variety of sampling policies e.g., uniform balanced, min rehearsal, max loss, min margin, min logit-distance, and min confidence \citep{chaudhry2018efficient,harun2023online,prabhu2023gdumb} and (4) buffer maintenance policy, with sample overwriting a randomly selected sample from the buffer once it is full, or with a class-based reservoir scheme \citep{de2021continual}, which approximates the full data distribution more closely than a uniform random selection.

\subsection{Experimental setup}

The model consists of two convolutional layers. These layers are responsible for scanning the image-converted time series data and constructing important features related to a potential anomaly. Each of these convolutional layers applies a set of filters, which can be thought of as small sliding windows that move over the image to detect different patterns \cite{Krizhevsky2017}. The first convolutional layer uses 80 filters with a kernel size of 3x3 pixels, meaning it applies 80 different filters to the image, each looking for a unique type of feature. The second layer also applies 80 filters with the same kernel size, but focuses on detecting more abstract patterns, building on the features identified in the earlier layers. By having multiple layers, the network can detect not just simple edges or textures, but also more complex patterns that correspond to specific types of wheel damage \cite{Zeiler2014}. These abstract features extracted from the preceding layers are subsequently fed into a fully connected layer, which executes the terminal classification.

\begin{table}[ht]
    \centering
    \caption{Layers specification}
    \label{tab:layerinfo}
    \resizebox{0.5\textwidth}{!}{%
    \begin{tabular}{c c c c c}
        \toprule
        \textbf{Layer} & \textbf{Filters} & \textbf{Kernel} & \textbf{Activation} & \textbf{Operations} \\
        \midrule
        \textbf{Conv. 1} & 80 & 3x3 & ReLU & MaxPool, Batch Norm. \\
        \textbf{Conv. 2} & 80 & 3x3 & ReLU & MaxPool, Batch Norm. \\
        \textbf{Dense} & - & - & Softmax & - \\
        \bottomrule
    \end{tabular}%
    }
\end{table}

\textbf{Operations.} To ensure that the learning process is stable and effective, the model employs batch normalization after each convolutional layer. Batch normalization is a technique that helps speed up training by normalizing the output of these layers, preventing the model from becoming unstable during learning \cite{Ioffe2015}. Following each convolutional layer, the model applies a max pooling operation. Max pooling is a down-sampling strategy that minimizes the dimensionality of the data by preserving only important features. This procedure is essential because it alleviates the computational burden on the system, thereby enhancing the model's efficiency without compromising vital information regarding potential damage. For example, after recognizing a crack in the wheel within a specific image region, max pooling guarantees the retention of this critical information while discarding superfluous data \cite{Rumelhart2013}. This information is summarized in Table~\ref{tab:layerinfo}.

\section{Wheel-rail simulation}
\label{chap:Chapter3}

This section describes the modeling and simulation process used to obtain the multi-domain data that mirrors real-world wheel-rail interaction across various operational scenarios, while inducing minor imperfections, such as flats and polygonization. Subsequently, to adjust to the challenging setting of continual learning, the simulation of environmental and operational conditions was organized in a sequence that pertains to the natural distribution of speeds, loads, and train types, across a calendar year. Moreover, to represent a challenging setting of wheel out-of-roundness detection with parsimonious sensorization, the simulation was captured with a single accelerometer and strain gauge positioned along the track.

\begin{figure}[ht]
    \centering
    \includegraphics[width=0.5\textwidth]{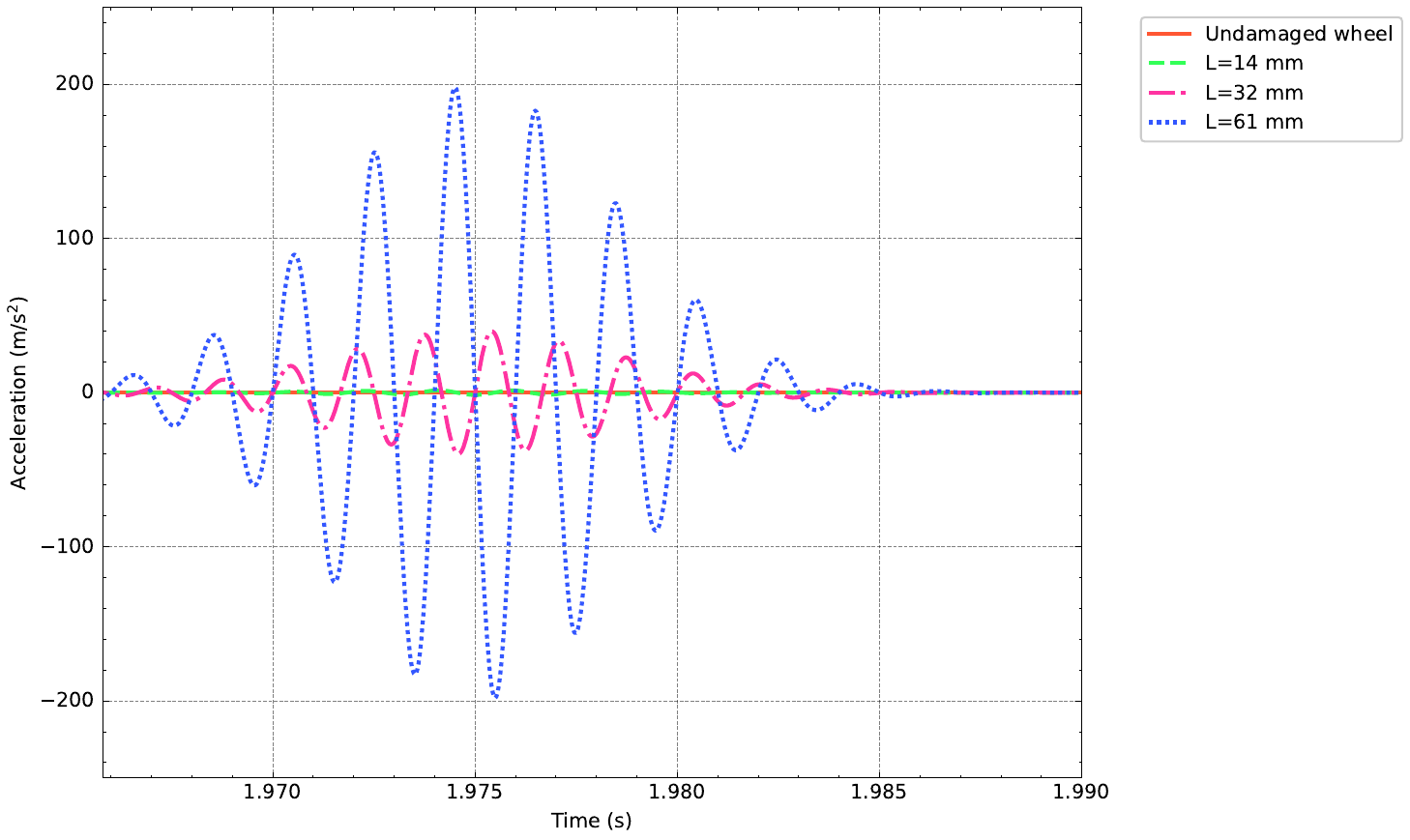}
    \caption{Acceleration for different flat lengths}
    \label{fig:acc_flats}
\end{figure}

\textbf{Flats.} For wheel flats, two flat length intervals (\(L_w\)) were considered, designated as L1 and L2. The uniform distributions U (25, 50) mm and U (50,100) mm define the lower and upper limits of the flat length of the wheel for each interval L1 and L2, respectively. The wheel flat depth (\(D_w\)) is calculated on the basis of equation \cite{Zhai2001}: \(D_w = \frac{L_w^2}{16R_w}\), in which \(R_w\) is the radius of the wheel. The vertical profile deviation of the wheel flat is defined as:
\begin{equation}
\scalebox{0.8}{$
\begin{aligned}
    Z = -\frac{D_w}{2} \left( 1 - \cos \frac{2\pi x_w}{L_w} \right) \cdot 
    H\Big(x_w - \big(2\pi R_w - L_w\big)\Big),  0 \leq x \leq 2\pi R
\end{aligned}
$}
\end{equation}

where \(H\) represents the Heaviside function and \(x_w\) is the coordinate aligned with the longitudinal direction of the track. Figure~\ref{fig:acc_flats} illustrates the effect of different severities of wheels flat.

\textbf{Polygonization.} For wheel polygonization, the periodic irregularity of the radial tread around the circumference of the wheel was considered by varying the wavelengths (\(\lambda\)) as a function of harmonic order (\(\theta\)) and wheel radius.
\begin{equation}
    \lambda = \frac{2\pi R_w}{\theta}, \theta = 1, 2, 3 \cdots, n
\end{equation}

\begin{figure}[ht]
    \centering
    \vspace{-10pt}
    \includegraphics[width=0.47\textwidth]{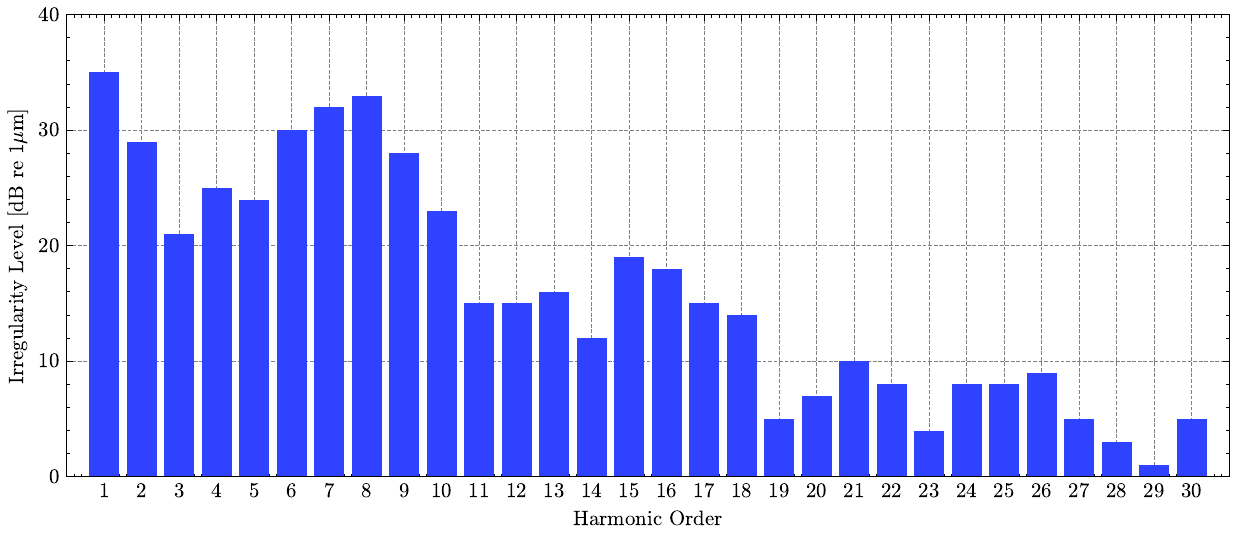}
    \caption{Amplitude (\(L_w\)) and harmonic (\(\theta\))}
    \label{fig:harmonic_order}
\end{figure}

The selected wheel profiles were characterized by the wavelengths in the first 20 harmonics, with the sixth to eighth harmonic orders being dominant, and different irregularity wheel profiles generated based on the sum of sine functions (\(H=20\)) as follows:
\begin{equation}
    w(x_w) = \sum_{\theta=1}^H A_\theta \sin \left( \frac{2\pi}{\lambda} x_w + \varphi_\theta \right),
\end{equation}
where \(A_\theta\) is the amplitude of the sine function for each wavelength, which is calculated by the function:
\begin{equation}
    A_{\theta} = \sqrt{2} \cdot 10^{\frac{L_{w}}{20}} \cdot w_{\text{ref}},
\end{equation}
with \(w_{ref} = 1\mu m\). The levels of wheel irregularity (\(L_w\)) were selected based on the irregularity spectrum in Figure~\ref{fig:harmonic_order}, produced with the measurement values of four wheels with polygonal damage \cite{Cai2019}. Taking into account the phase angles to the sine functions that are uniformly and randomly distributed between \(0\) and \(2\pi\), several irregularity profiles of the wheel were generated to obtain different damage severities between 0.8 and 1.2 mm.

\begin{figure}[ht]
    \centering
    \includegraphics[width=0.5\textwidth]{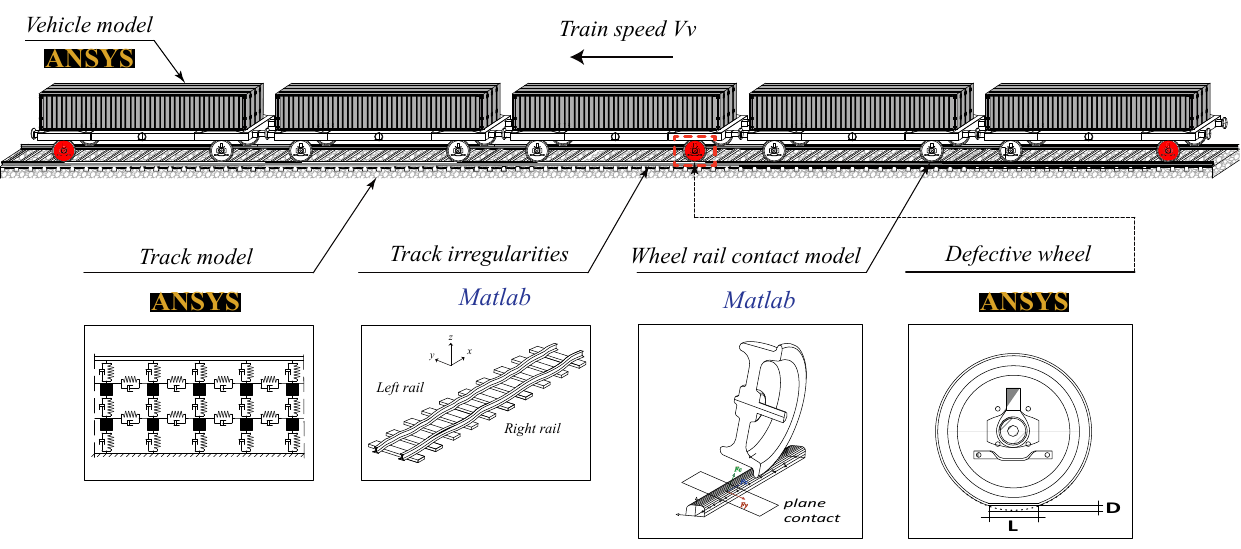}
    \caption{Train-track dynamic interaction}
    \label{fig:train_track_dynamic}
\end{figure}

\textbf{Train-track dynamic interaction.} Different modes of interaction were numerically investigated using VSI software, whose validation and detailed description are documented in \cite{Montenegro2015}. The VSI model integrates the train and the track through a three-dimensional wheel-rail contact model that uses Hertzian theory to calculate normal contact forces and the USETAB routine to determine the tangential forces arising from rolling friction creep \cite{Hertz1882, kalker1996book}. Designed in MATLAB®, the VSI software incorporates structural matrices from both the vehicle and the track, which have been independently modeled using finite element (FE) analysis. In particular, the track was characterized using beam elements to represent the rails and sleeping areas. Spring-dashpot elements were used to simulate the behavior of the flexible layers, including the ballast and fasteners/pads, while mass-point elements were used to account for the ballast's mass. The train model was developed in ANSYS® via a multibody framework. This involved using spring-dashpot elements to mimic the flexibility of both primary and secondary suspensions, rigid beams to represent the vehicle's rigid body motions, and mass point elements positioned at the center of gravity of each component, such as the carbody, bogies, and wheel sets, to capture their mass and inertial properties. Figure~\ref{fig:train_track_dynamic} provides a graphical representation of the numerical model.

\textbf{Track conditions.} In real conditions, the rails also present small imperfections that can significantly affect the values of the wheel-rail contact force. In this context, 10 irregularity profiles were generated for wavelengths between 1m and 30m, covering the D1 wavelength interval defined by the EN 13848-2 standard \cite{standard2003railway}, with a sampling discretization of 1mm. The amplitude of the irregularity profile was varied between -2mm and 2mm, for a total simulation length of 100m. These were based on real track irregularities of the Northern Line of the Portuguese Railway Network, measured with a track inspection vehicle EM120 every six months. It is important to note that the wavelengths used to generate these track irregularities are significantly longer than the wheel flat and polygonization. Thus, the exit frequencies due to the track are much lower than those of a defective wheel. More details on the generation of unevenness profiles can be found elsewhere \cite{Mosleh2020}.

\textbf{Train types and speed.} The simulated data consists of two types of trains: the Laagrss vehicle and the Alfa vehicle. The former is designed for hauling freight, typically carrying heavier loads at slower speed. In contrast, Alfa is a train built for passenger transport, prioritizing speed and lighter weight. For Laagrss, the configurations include passages in the range 40 to 120 km/h, and for Alfa in the range 40 to 220 km/h.

\begin{figure}[ht]
    \centering
    \begin{subfigure}{0.23\textwidth}
        \includegraphics[width=\linewidth]{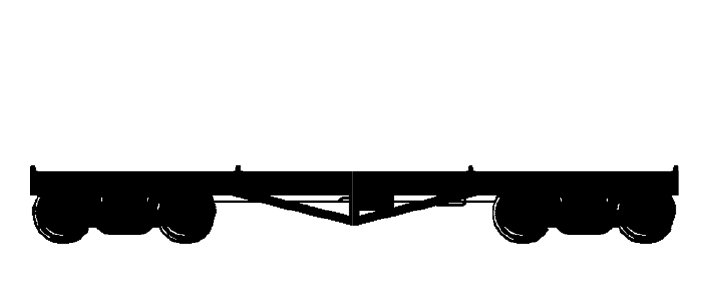}
        \caption{Empty}
    \end{subfigure}\hfill
    \begin{subfigure}{0.23\textwidth}
        \includegraphics[width=\linewidth]{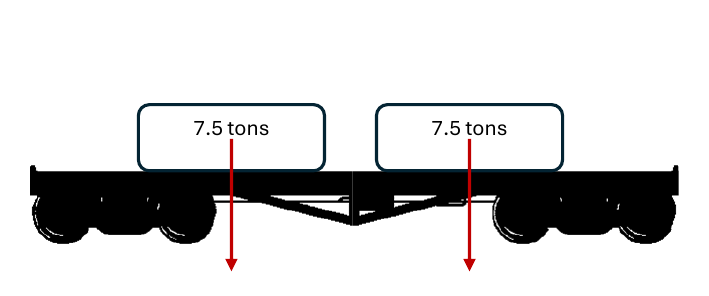}
        \caption{Half}
    \end{subfigure}\hfill
    \begin{subfigure}{0.23\textwidth}
        \includegraphics[width=\linewidth]{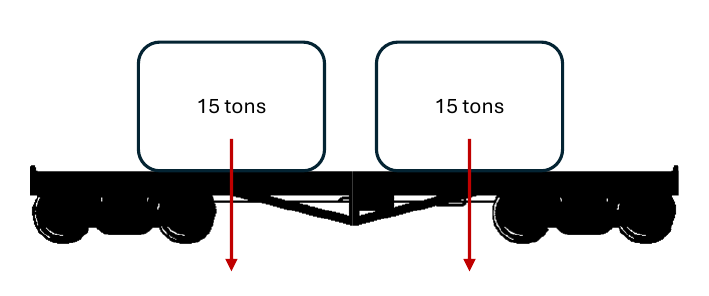}
        \caption{Full}
        \label{fig:fullscheme}
    \end{subfigure}\hfill
    \begin{subfigure}{0.23\textwidth}
        \includegraphics[width=\linewidth]{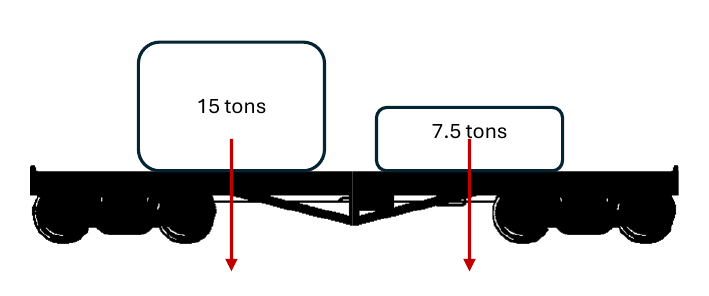}
        \caption{Unbalance 1}
    \end{subfigure}\hfill
    \begin{subfigure}{0.23\textwidth}
        \includegraphics[width=\linewidth]{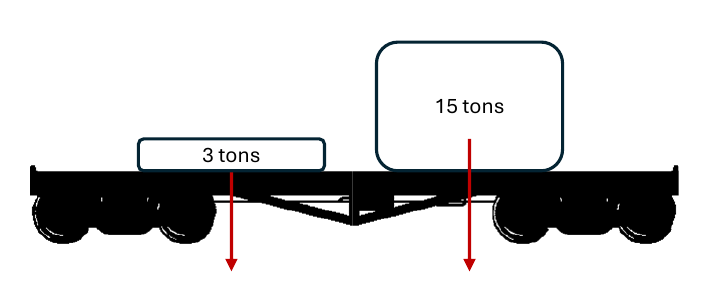}
        \caption{Unbalance 2}
    \end{subfigure}\hfill
    \begin{subfigure}{0.23\textwidth}
        \includegraphics[width=\linewidth]{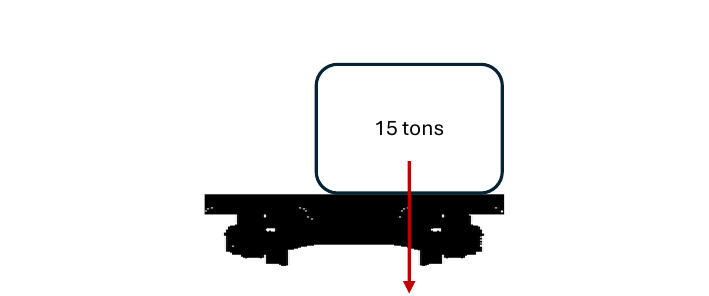}
        \caption{Unbalance 3}
    \end{subfigure}

    \caption{Load schemes}
    \label{fig:trainschemes}
\end{figure}

\textbf{Load.} Figure~\ref{fig:trainschemes} illustrates the various train load schemes considered. The configurations include: (a) an empty train, where there is no load; (b) a half-loaded train with an equal load distribution of 7.5 tons; and (c) a fully loaded train with equal load distribution, applying 15 tons. Furthermore, three unbalanced load configurations are depicted: (d) unbalance 1, where one side carries 15 tons while the other side carries 7.5 tons, forcing more on one of the sides, increasing the stress on those wheels; (e) unbalance 2, where one side carries 15 tons and the other side carries a smaller load of 3 tons; and (f) unbalance 3, where the entire load of 15 tons is concentrated on one side.

\subsection{Multi-domain data}

Table \ref{tab:comparision_baseline_damage} provides a detailed comparison between the baseline and damage domains, resulting of combinatorially exploring the aforementioned EOVs. Baseline domains have six different train load configurations, including full, half, empty, and three unbalanced load scenarios, with speeds ranging from 40 to 220 km/h and varying irregularity profiles (1-4). In contrast, damage domains focus on specific defect scenarios: flat wheels and polygonized wheels. Both damage domains are characterized by a fully loaded configuration, limited speed ranges (60, 80, 100, 120 and 200 km/h), and specific defect locations on the Laagrss and Alfa trains. The amplitude and depth of defects, as well as the harmonic orders, are outlined for each damage type, with detailed metrics for flat and polygonized wheels.

\begin{table*}[ht]
    \centering
    \caption{EOVs for domain division}
    \label{tab:comparision_baseline_damage}
    \begin{tabular}{l c c c} 
    \toprule
         & \textbf{Baseline} & \textbf{Flat} & \textbf{Polygonized} \\ 
    \midrule
        \textbf{Train load} & 6 (full, half, empty, and 3 diff. unbalanced) & 1 (full) & 1 (full) \\
        \textbf{Irregularity profiles} & 96 & 24 & 24 \\
        \textbf{Train speeds (km/h)} & 40-220 & 60-200 & 60-200 \\
        \textbf{Defect locations} & - & 3rd wagon, 1st/3rd left & 1st wagon, 1st right \\
        \textbf{Amplitude (mm)} & - & \makecell[c]{10-20, 25-50, 40-100} & \makecell[c]{.25-.35, 0.55-.75} \\
        \textbf{Depth (mm)} & - & \makecell[c]{.02-.06, .09-.16, .23-.36} & \makecell[c]{6-8, 12-14, 17-20, 29-30} \\
    \bottomrule
    \end{tabular}
\end{table*}

\begin{figure}[!ht]
    \centering
    \begin{subfigure}[b]{\columnwidth} 
        \centering
        \includegraphics[width=0.9\linewidth]{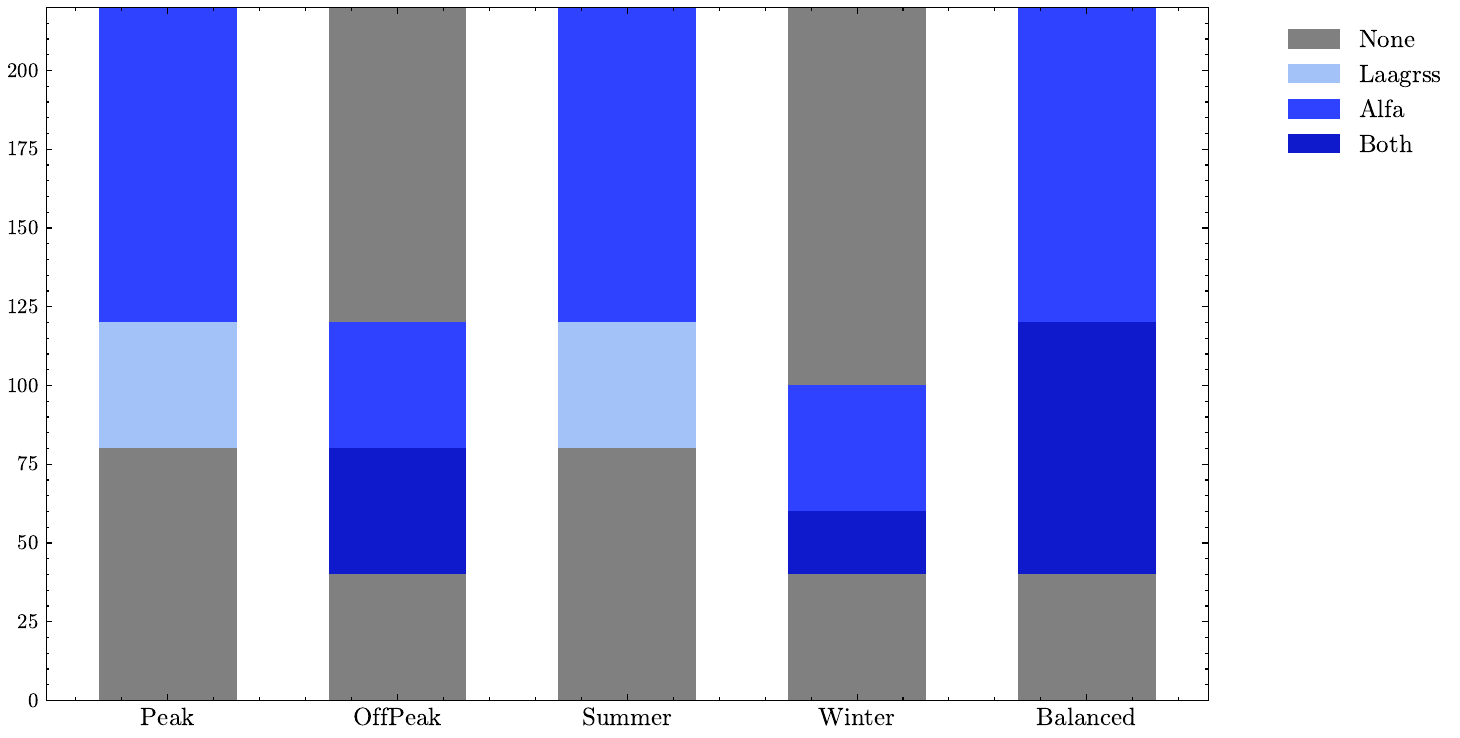}
        \caption{Speed (in km/h)}
        \label{fig:train_speed}
    \end{subfigure}
    \hfill
    \begin{subfigure}[b]{\columnwidth} 
        \centering
        \includegraphics[width=0.9\linewidth]{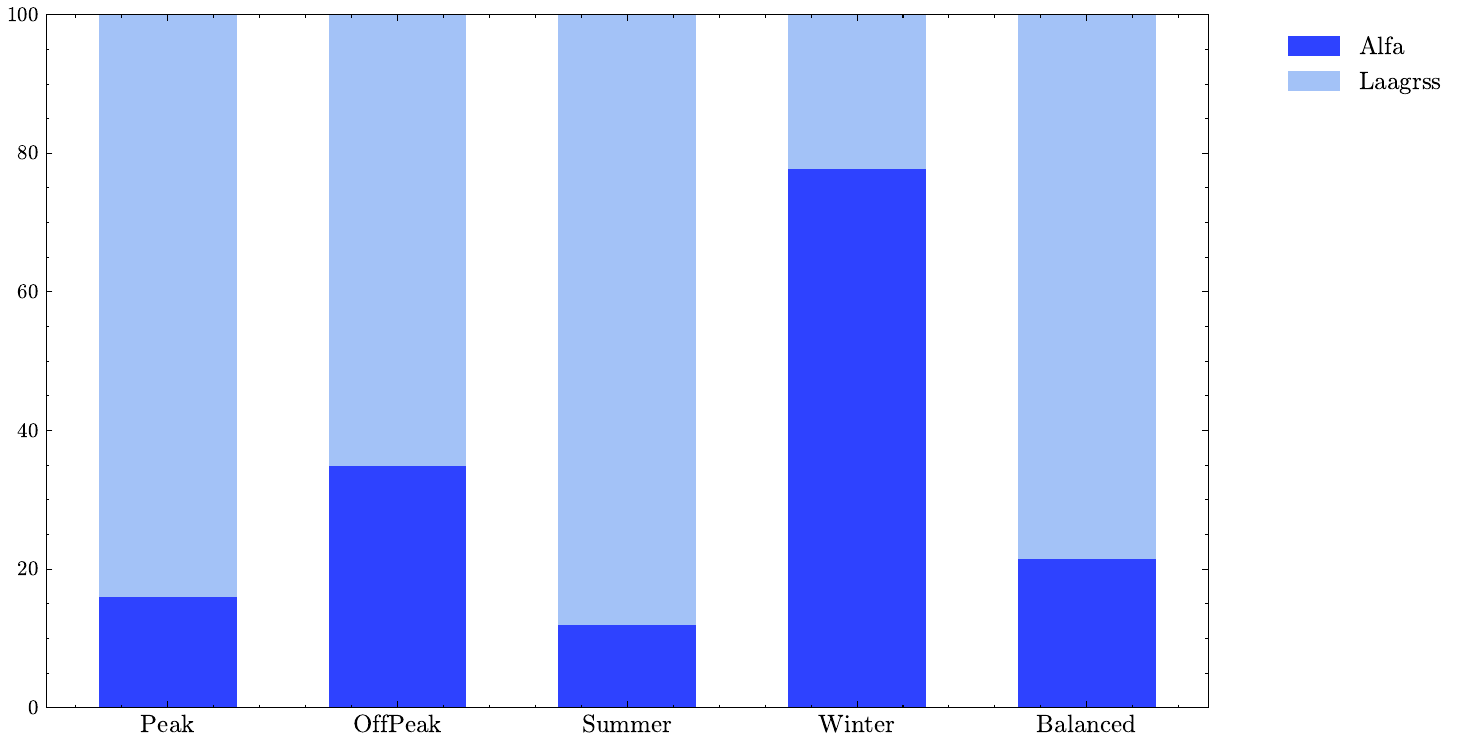}
        \caption{Train type}
        \label{fig:train_percentage}
    \end{subfigure}
    \begin{subfigure}[b]{\columnwidth} 
        \centering
        \includegraphics[width=0.9\linewidth]{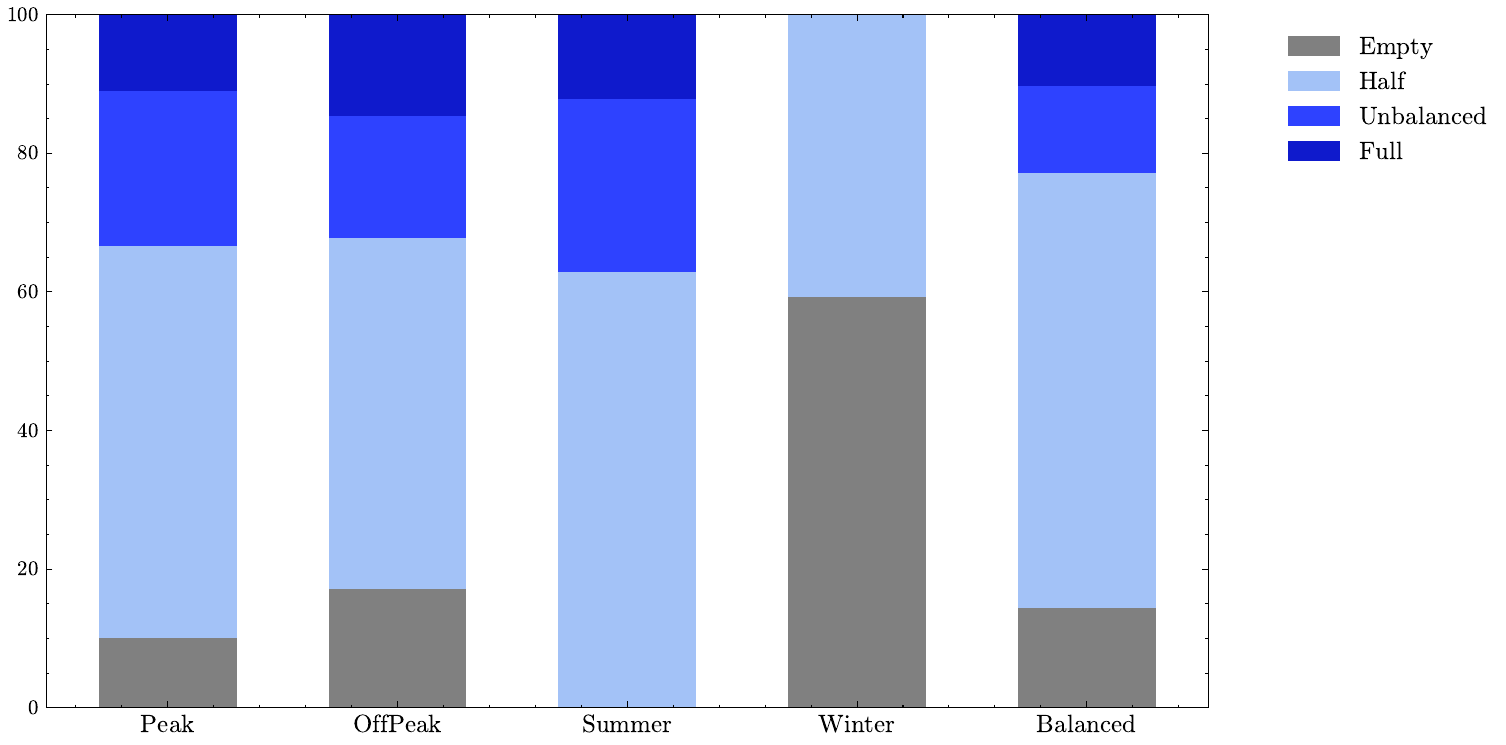}
        \caption{Cargo load}
        \label{fig:load_percentage}
    \end{subfigure}
    
    \caption{Distribution of EOVs across different train types}
    \label{fig:sample_train_partitioning}
\end{figure}

\textbf{Natural sequence.} However, to truly integrate aspects of continual learning in real-world applications, and evaluate if the model can retain previous knowledge, while managing new information across various data domains, the training procedure must reflect potential temporal evolutions that honor the natural-time sequence. Therefore, several domains were carefully partitioned and organized in order to pertain both seasonal and recurrent patterns. The simulated domains, illustrated in Figure~\ref{fig:sample_train_partitioning}, are:
\begin{itemize}
    \item \textbf{Peak}: Comprise of higher speed trains, minimizing traffic disruptions during peak seasons.
    \item \textbf{Off peak}: Features slower speed trains operating in a less congested railway setting, typical of off-peak seasons.
    \item \textbf{Summer boom}: High flow of commerce and transport, characterized by high-speed trains with fully loaded wagons.
    \item \textbf{Winter bust}: Featuring the slowest speeds and never completely filled wagons, indicative of a slow flow of transport and goods.
    \item \textbf{Balanced}: Operating at medium speeds, optimizing for fuel efficiency and travel time, mantaining a steady flow of traffic.
\end{itemize}

\section{Experimental results}
\label{chap:Chapter4}

This section presents an analysis of BOLT-RM's performance. The results demonstrate the impact of continual learning and hyperparameter tuning on the accuracy and robustness of the model. The evaluation includes the model’s ability to handle different train-track interactions, loads, and speeds, while highlighting three common learning metrics: forward transfer, backward transfer, and domain-specific performance.

\textbf{Domain-specific performance.} The average domain accuracy (ACC) evaluates the overall performance of the model in all domains. Higher values indicate better final performance across all domains. It is calculated as:
\begin{equation}
    \text{ACC} = \frac{1}{D} \sum_{i=1}^{D} a_{D,i},
\end{equation}
where \(a_{D,i}\) represents the accuracy on the \(i^{th}\) domain after training all the \(D\) domains.

\textbf{Forward transfer.} The learning accuracy (LA) evaluates the model's ability to learn new domains by using prior knowledge. Higher values indicate better learning transfer across domains. It is calculated as:
\begin{equation}
    \text{LA} = \frac{1}{D} \sum_{i=1}^{D} a_{i,i},
\end{equation}
where \(a_{i,i}\) represents the accuracy immediately after training in the \(i^{th}\) domain.

\textbf{Backward transfer.} The forgetting measure (FM) measures how much the model has forgotten previous domains after learning new domains. Lower values are better because they indicate that the model retains more knowledge from previous domains, with negative values indicating positive backward transfer. It is calculated as:
\begin{equation}
    \text{FM} = \frac{1}{D - 1} \sum_{i=1}^{D-1} \max_{l \in \{1, \ldots, D-1\}} (a_{l,i} - a_{D,i})
\end{equation}
where \(a_{l,i}\) represents the accuracy in the \(i^{th}\) domain after learning the \(l^{th}\) domain, and \(a_{D,i}\) represents the accuracy in the \(i^{th}\) domain after learning all domains.

\begin{figure}[ht]
    \centering
    \includegraphics[width=0.4\textwidth]{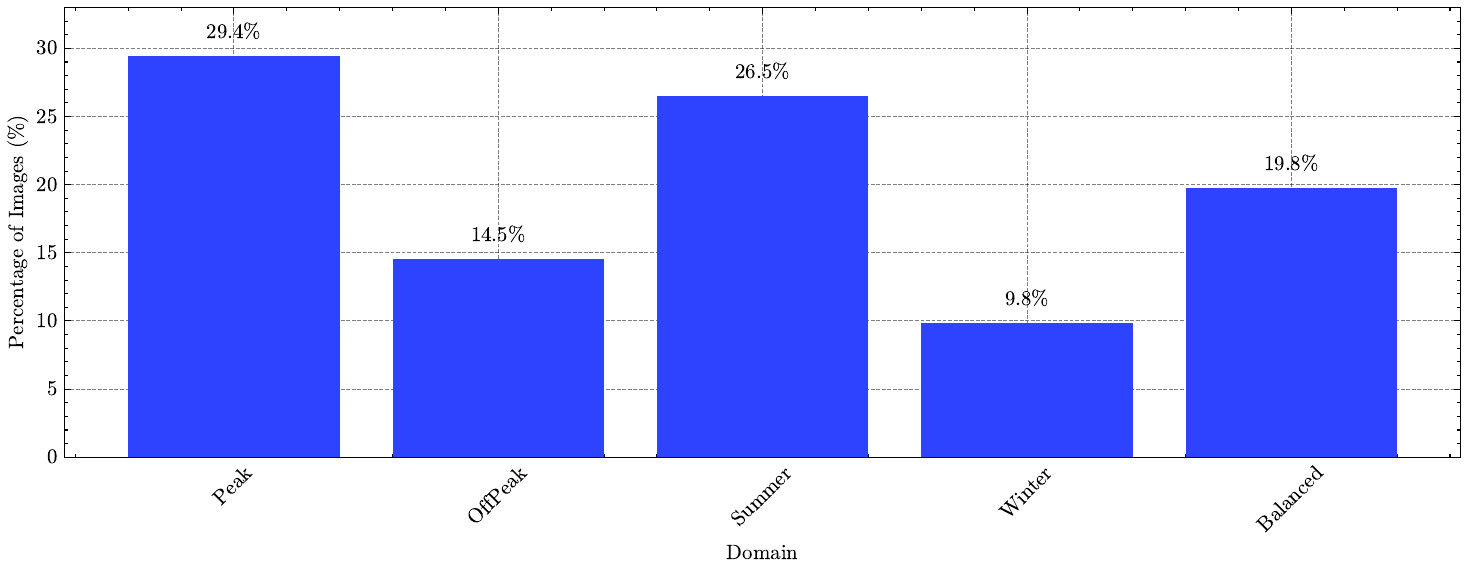}
    \caption{Images per domain}
    \label{fig:imagesperdomain}
\end{figure}

\textbf{Data distribution.} Figure~\ref{fig:imagesperdomain} shows the number of MTF images per domain used for training and testing purposes. This preprocessing step, which transformed the raw signals into structured MTF images, is beneficial in enhancing the model's ability to detect patterns related to wheel anomalies.

\begin{figure}[ht]
    \centering
    \includegraphics[width=0.5\textwidth]{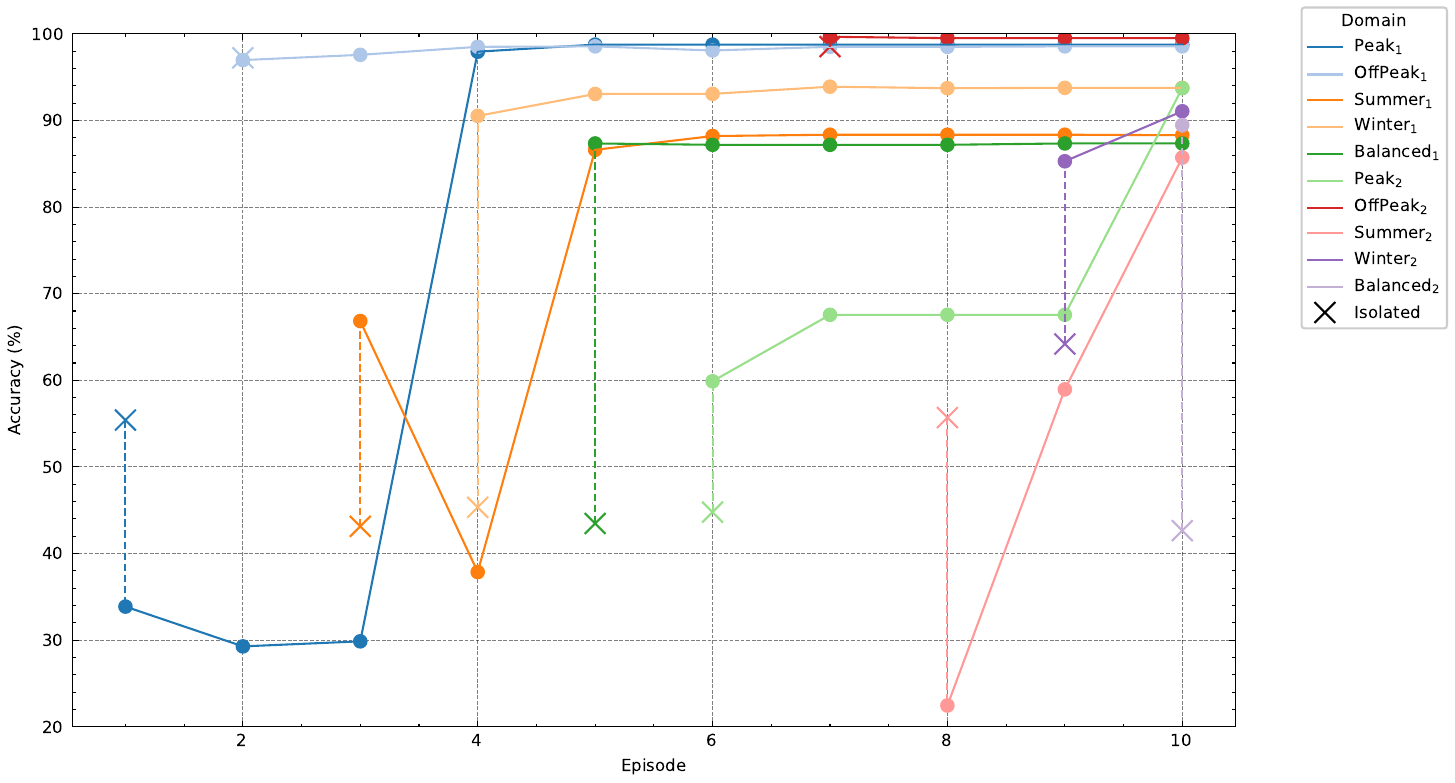}
    \caption{Evolution of domain accuracy}
    \label{fig:plotresults}
\end{figure}

\textbf{Isolated ablation.} Figure~\ref{fig:plotresults} shows the evolution of the accuracy of the domain in episodes, comparing BOLT-RM with the isolated model. The isolated model is a simple continual learning model that does not share information between domains, so it only trains the current data and ignores the past data. The results highlight the ability of BOLT-RM to improve its accuracy over time, even in domains where initial performance may have been lower. Although the isolated model, represented by crosses, occasionally outperforms BOLT-RM in earlier episodes, BOLT-RM demonstrates an adaptive capacity to improve, eventually surpassing the isolated model as the episodes progress. Such gradual improvement highlights the strength of BOLT-RM in learning from multiple domains and using knowledge sharing, even when some domains initially pose greater challenges. Beyond accuracy, continual learning setting requires further inspection on the aforementioned three performance metrics. Table~\ref{tab:metrics} summarizes these metrics, underscoring the benefits of BOLT-RM over the isolated model. BOLT-RM considerably exceeds the isolated model in the average domain accuracy, achieving 0.93 versus 0.54. This indicates the ability in BOLT-RM to use prior domain knowledge to improve overall performance. The forward transfer score further underscores this benefit, with BOLT-RM scoring 0.73, suggesting better knowledge transfer. Moreover, BOLT-RM exhibits minimal forgetting ($3.47 \times 10^{-4}$), highlighting its ability to retain learned information, unlike the isolated model, which shows no backward transfer. 

\begin{table}[!ht]
    \caption{Metrics for BOLT-RM vs. Isolated model}
    \label{tab:metrics}
    \resizebox{\columnwidth}{!}{%
    \begin{tabular}{c c c}
    \toprule
        \textbf{Metric} & \textbf{BOLT-RM} & \textbf{Isolated}\\
    \midrule
        Average Domain Accuracy & 0.93 & 0.54\\
        Learning Accuracy (Forward Transfer) & 0.73 & 0.54 \\
        Forgetting (Backward Transfer) & $3.47 \times 10^{-4}$ & 0\\
    \bottomrule
    \end{tabular}%
    }
\end{table}

\begin{figure}[ht]
    \centering
    \includegraphics[width=0.5\textwidth]{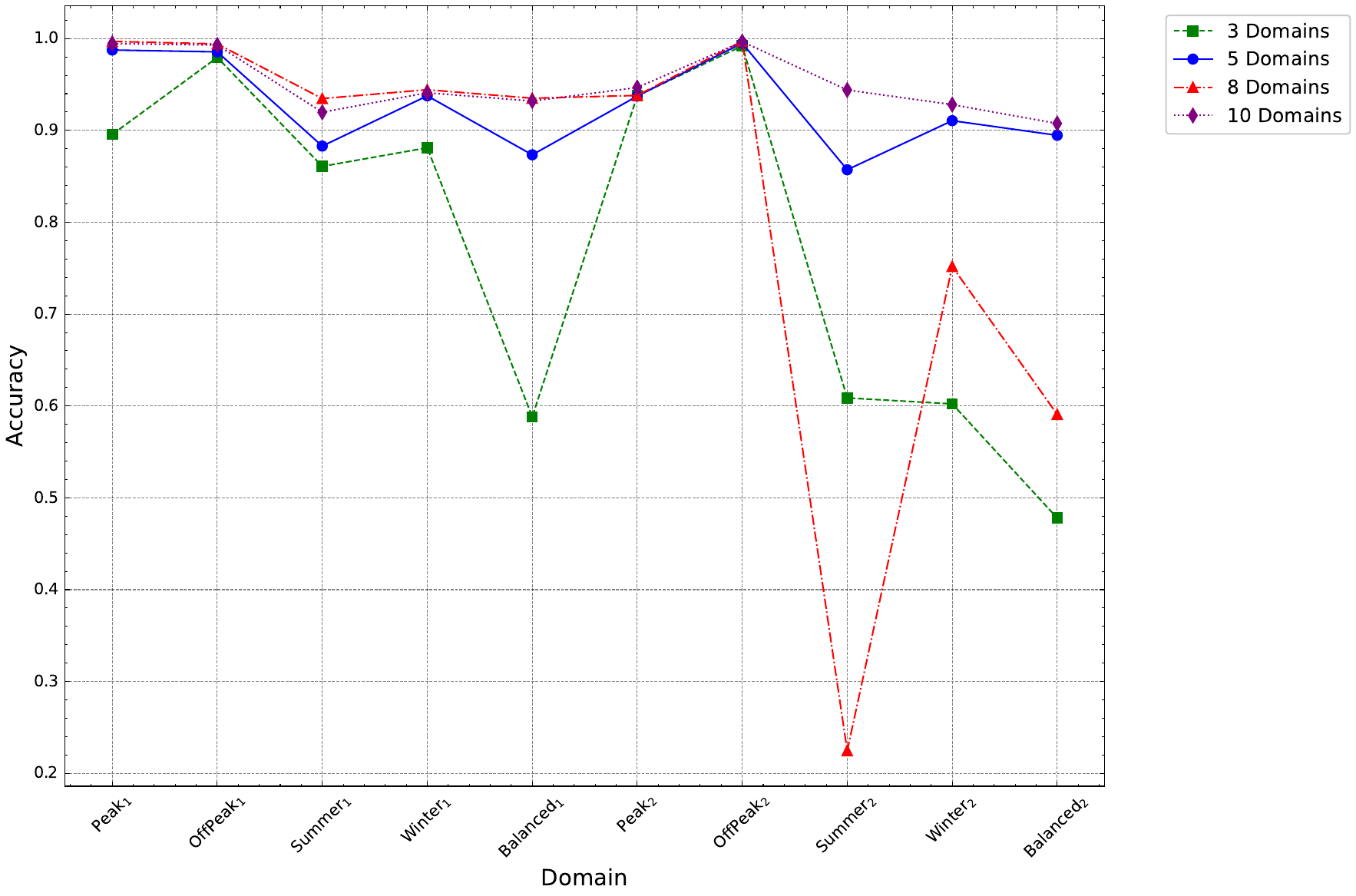}
    \caption{Final accuracy per domain for different maximum numbers of domains}
    \label{fig:avgdiffdomains}
\end{figure}

\textbf{ER size ablation.} Table~\ref{tab:modelcompardomain} presents the experiments conducted to vary the maximum number of domains included in the experience replay (ER), which differs from the default configuration of using five domains and the selection criteria for the ER inspired by boosting to take the lowest performance domains for it. The total number of domains was limited for a maximum of 3, 5, 8, or 10 domains. Figure~\ref{fig:avgdiffdomains} present the final accuracy for each domain in the four configurations. Among these, the use with 10 domains achieved the highest overall accuracy, which is expected, since it uses all the past data. This setup also demonstrated superior forward transfer (95\%), likely due to the substantial amount of information available for new data, allowing more effective learning. However, the 10-domain configuration did not achieve the best backward transfer (\(1.96 \times 10^{-3}\)). Instead, the default configuration for the 5-domains outperformed it (\(3.47 \times 10^{-4}\)). Using 10 domains can introduce excessive data diversity and potential ``overload'', preventing improvements in previously learned domains compared to the more focused configuration of five domains. Hence, the 5-domain configuration remains a favorable compromise, offering significantly faster training while maintaining a high average accuracy (92.6\% vs. 95.0\%), forward transfer (73.2\% vs. 76.2\%) only slightly lower than the 10-domain configuration, and considerably better backward transfer (\(3.47 \times 10^{-4}\) vs. \(1.96 \times 10^{-3}\)).   An 8-domain configuration was also evaluated. Although it showed a slight improvement in forward transfer (76.1\%) compared to 5 domains (73.2\%), the difference was minimal. Moreover, the 8-domain setup suffered from catastrophic forgetting in domain 8 during the final training episode, underscoring that increasing the number of domains in ER does not necessarily enhance performance. In contrast, using only 3 domains in the ER resulted in the poorest overall performance (78.2\%), with stagnation in domain accuracies after a certain point and the lowest forward transfer (61.7\%). This shows that restricting ER to too few domains limits both new learning and the improvement of previously learned domains. Although this setup shows no backward transfer (0), which means that there is no forgetting, this was primarily because the model stopped improving entirely, leaving its performance on a plateau.

\begin{table}[ht]
    \centering
    \caption{Metrics for BOLT-RM vs. Isolated model}
    \label{tab:modelcompardomain}
    \resizebox{\columnwidth}{!}{%
    \begin{tabular}{c c c c c}
    \toprule
        \textbf{Metric} & \textbf{3 Domains} & \textbf{5 Domains} & \textbf{8 Domains} & \textbf{10 Domains}\\
    \midrule
        ACC & 0.78 & \textbf{0.93} & 0.83 & 0.95\\
        LA & 0.62 & \textbf{0.73} & 0.76 & 0.76\\
        FM & 0 & \(\boldsymbol{3.47 \times 10^{-4}}\) & \(7.79 \times 10^{-2}\) & \(1.96 \times 10^{-3}\)\\
    \bottomrule
    \end{tabular}
    }
\end{table}

\begin{figure}[!ht]
    \centering
    \includegraphics[width=0.5\textwidth]{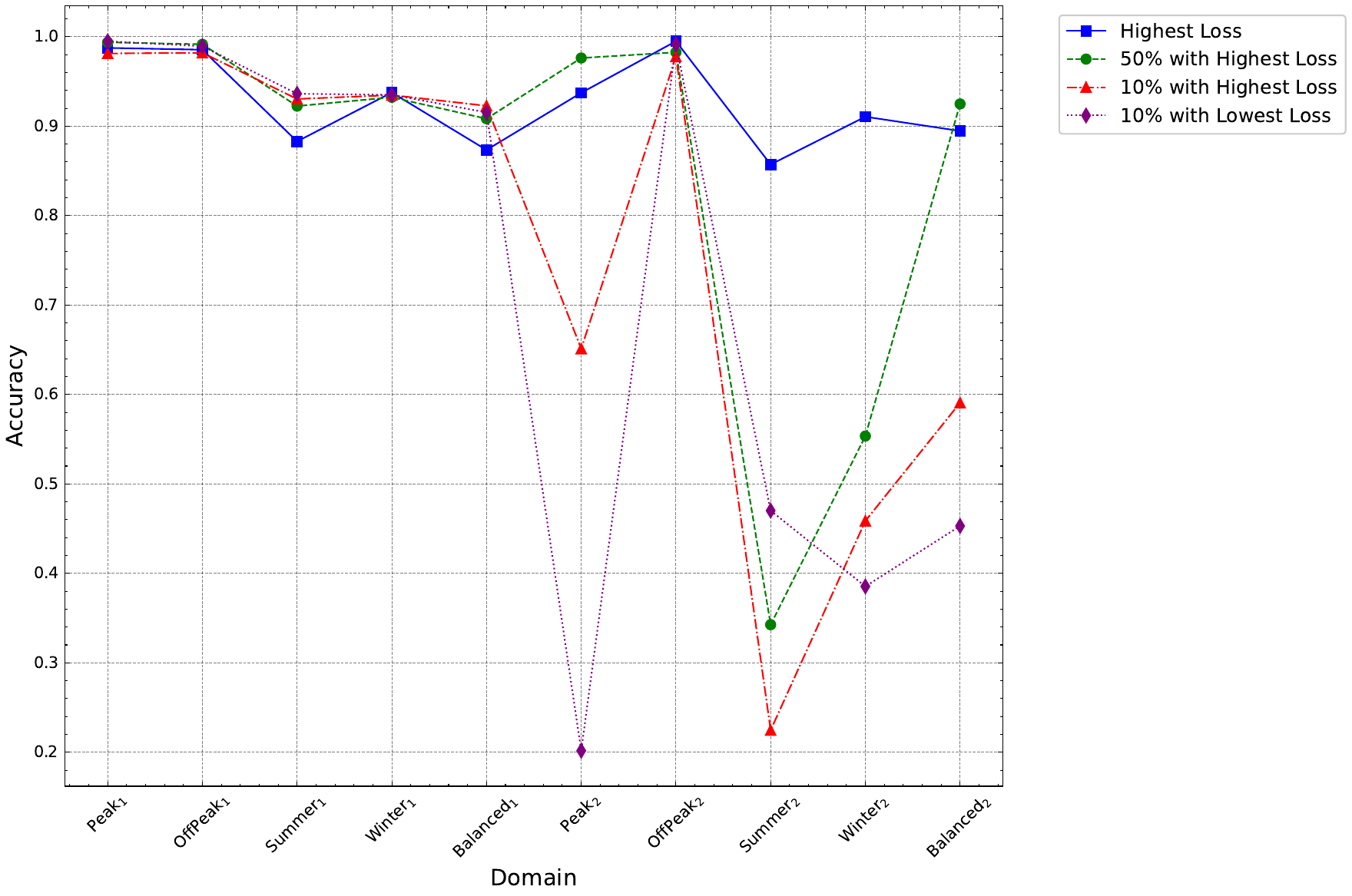}
    \caption{Final accuracy per domain for different domain selection criteria}
    \label{fig:avgdiffmetric}
\end{figure}

\textbf{ER selection ablation.} Figure~\ref{fig:avgdiffmetric} compares different selection criteria for including domains in ER, to examine whether including well-performing domains could improve forward and backward transfer. The default approach of always selecting highest-loss domains (boosting-inspired) was tested against using a 50/50 split between highest-loss and lowest-loss domains, and more extreme divisions such as 10\% highest-loss and 90\% lowest-loss or vice versa. However, this approach resulted in a decline in average accuracy (85.3\% vs. 92.6\%) and forward transfer (67.8\% vs. 73.2\%), with only a slight improvement in backward transfer (\(2.72 \times 10^{-4}\) vs. \(3.47 \times 10^{-4}\)). Thus, mixing high- and low-performing domains does not outweigh the benefits of a pure boosting-inspired strategy. Further experimentation with 10\%/90\% splits (in both directions) did not produce improvements. These configurations resulted in lower average accuracy (76.5\% and 72.7\% vs. 92.6\%), reduced forward transfer (59.7\% and 61.3\% vs. 73.2\%), and worse backward transfer (\(2.97 \times 10^{-3}\) and \(2.56 \times 10^{-3}\) vs. \(3.47 \times 10^{-4}\)). These results reaffirm that maintaining the boosting-inspired strategy, i.e. selecting domains with the highest loss, is preferable.

\begin{table}[ht]
    \centering
    \caption{Metrics for BOLT-RM vs. Isolated Model}
    \label{tab:modelcomparsplit}
    \resizebox{\columnwidth}{!}{%
    \begin{tabular}{c c c c c}
    \toprule
        \textbf{Metric} & \textbf{Highest} & \textbf{50/50 Split} & \textbf{10\% Highest} & \textbf{10\% Lowest}\\
    \midrule
        ACC & \textbf{0.93} & 0.85 & 0.77 & 0.73\\
        LA & \textbf{0.73} & 0.68 & 0.60 & 0.61\\
        FM & \(\boldsymbol{3.47 \times 10^{-4}}\) & \(2.72 \times 10^{-4}\) & \(2.97 \times 10^{-3}\) & \(2.56 \times 10^{-3}\)\\
    \bottomrule
    \end{tabular}
    }
\end{table}

\begin{figure}[ht]
    \centering
    \includegraphics[width=0.45\textwidth]{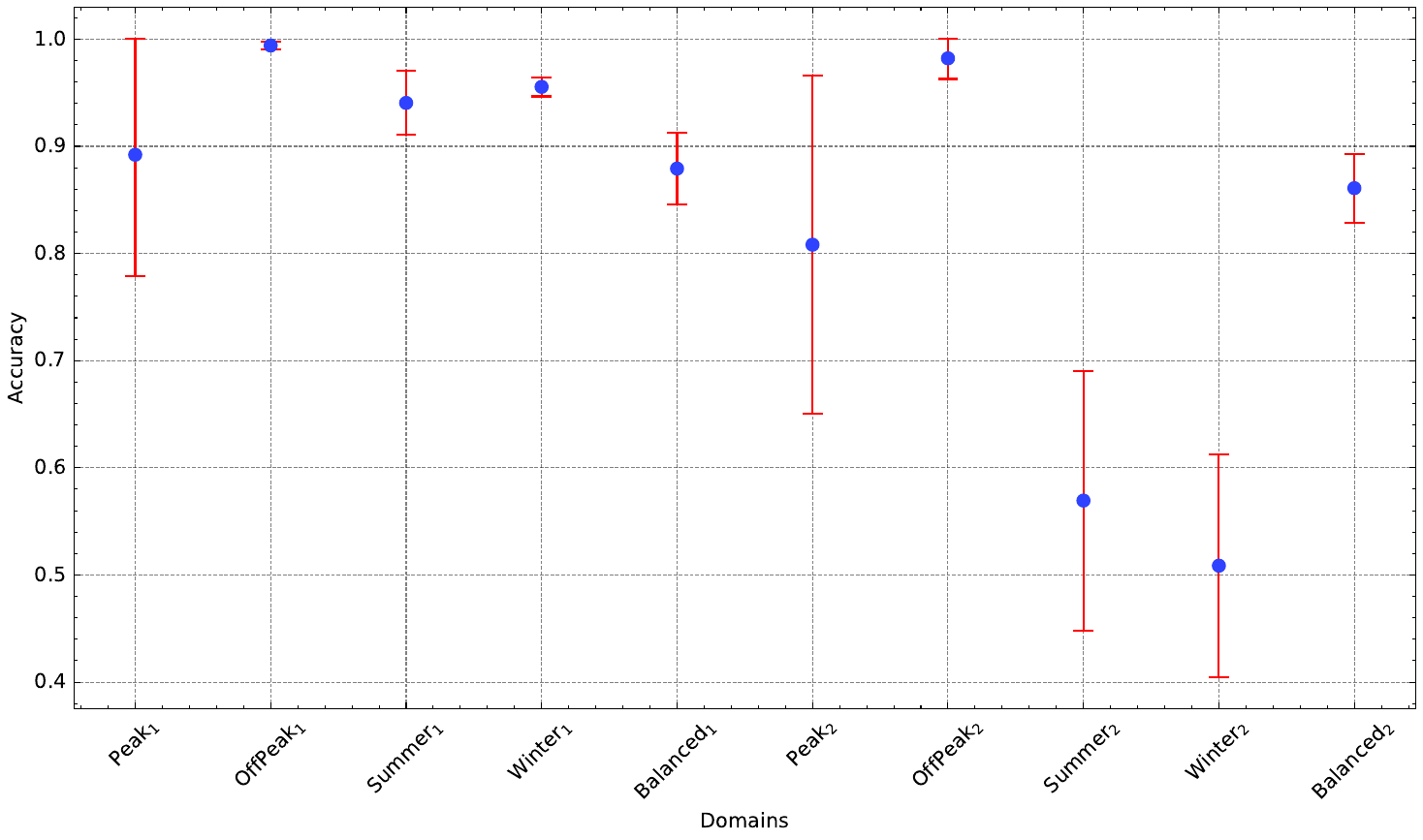}
    \caption{Domain accuracy with 95\% confidence intervals based on 10 runs}
    \label{fig:domain_error_acc}
\end{figure}

\textbf{Intra-domain data ablation.} To validate the impact of the order of data, within each domain, 10 different runs were performed. Figure~\ref{fig:domain_error_acc} shows the mean accuracies with error bars representing the confidence intervals, highlighting the stability and reliability of the model predictions across different domains, where each point represents the mean accuracy for a specific domain, with red error bars indicating the 95\% confidence interval. Although the model performed robustly in general, domains 8 and 9 initially showed the lowest accuracy and did not show any improvement over episodes in earlier runs. However, after hyperparameter tuning, these domains achieved accuracy levels comparable to those of other domains and demonstrated improvement over episodes. In particular, domains 3 and 8, and domains 4 and 9, share identical scenarios. The underperformance in earlier runs can be ascribed to several factors, one of them being that these domains are evaluated later in the training, potentially suffering from task interference. Moreover, despite operational similarities with domains 3 and 4, domains 8 and 9 have greater noise and variability, making generalization difficult. Single-pass training (one epoch per run) also hinders the model adaptation to these challenging scenarios. These results underscore the importance of fine-tuning hyperparameters to address limitations in subsequent episodes, where increased data variability may hinder performance.

\begin{figure}[ht]
    \centering
    \includegraphics[width=\columnwidth]{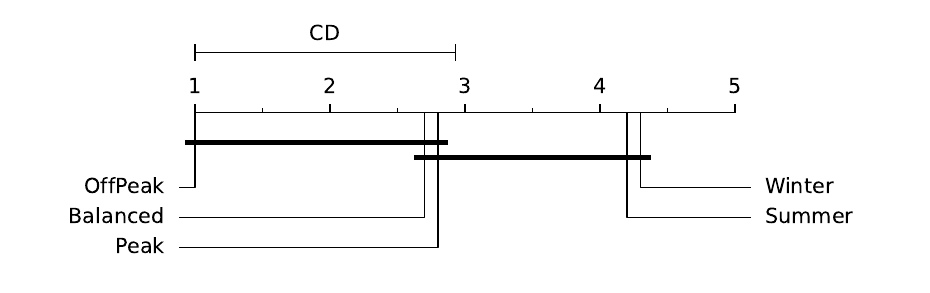}
    \caption{Critical difference diagram}
    \label{fig:cdd}
\end{figure}

\textbf{Inter-domain ablation.} To further assess the statistical significance of performance differences between domains, the Friedman test was applied, a nonparametric test designed to detect performance differences when the same models are evaluated under different conditions. The result, with a p-value of \(9.2 \times 10^{-4}\), indicates that there are statistically significant differences between the domains. To further investigate these disparities, the critical difference diagram, depicted in Figure~\ref{fig:cdd}, was used to detect notable differences and similarities between domains. The interconnected domains in the diagram suggest a lack of significant performance differences to be able to see the similarities between the domains. A lower rank on the diagram signifies superior domain accuracy. For example, domains 1 and 6, which represent high-speed trains during peak seasons, have a smaller difference, likely due to their comparable operational characteristics. In contrast, domains 3 and 8, characterized by fully loaded trains during the summer season, display noticeable differences in performance, with several other domains positioned between them in the critical difference diagram. This suggests that despite sharing operational conditions, such as fully loaded wagons and high-speed trains, other factors, such as noise levels, defect characteristics, or variability in train schedules, can contribute to observed differences. Similarly, the larger difference in performance of domains 4 and 9, indicative of slower speeds and reduced wagon loads in winter, may reflect the unique challenges these conditions pose to the model. These results underscore the importance of reflecting domain-specific characteristics in model design and evaluation.

\begin{figure}[ht]
    \centering
    \includegraphics[width=0.35\textwidth]{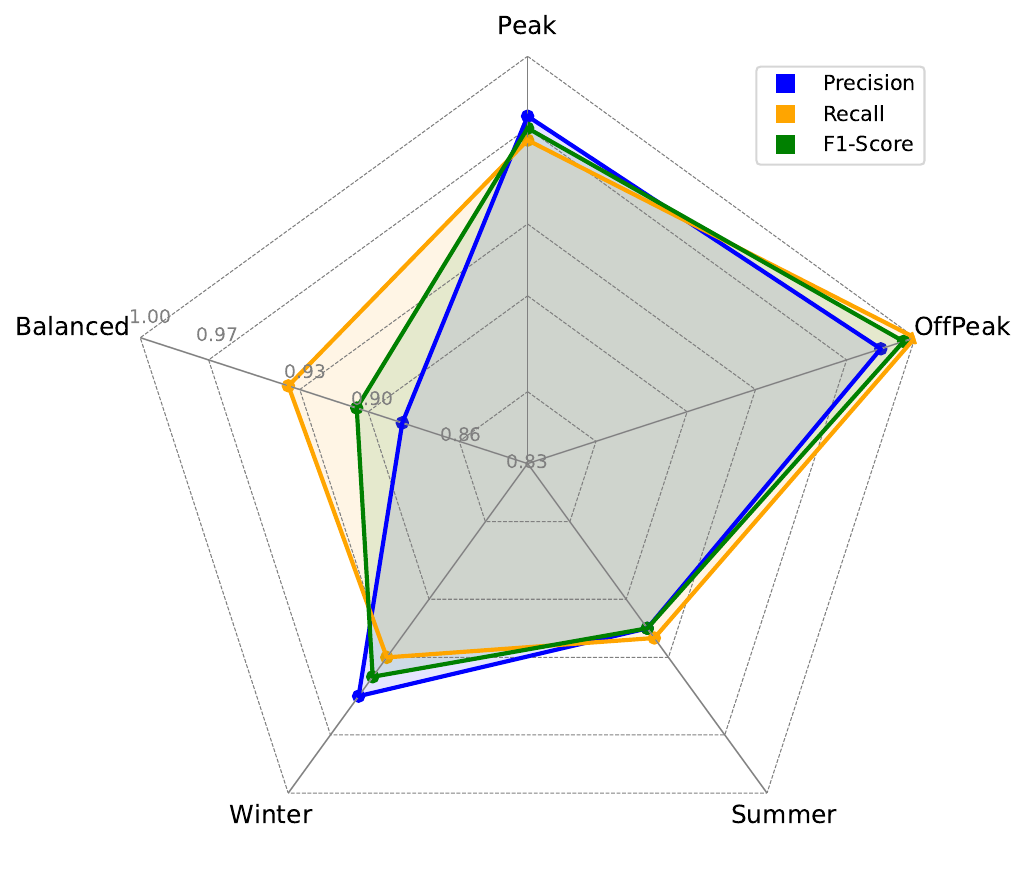}
    \caption{Performance metrics for each domain}
    \label{fig:spider_chart}
\end{figure}

Beyond accuracy, the imbalanced nature of data requires further inspection on other performance metrics. To complement overall accuracy, domain-specific performance metrics such as precision, recall, and the F1-score were calculated. Precision assesses the model's accuracy in identifying true positives among all predicted positives, recall measures its success in detecting all true positives, and the F1-score is the harmonic mean of precision and recall, providing a balanced performance metric. Table~\ref{fig:spider_chart} presents these metrics for each domain. The results indicate a near-perfect performance of the model in domains 1, 2 and 7 with F1-scores of 0.99 and 1.00, which means optimal fault detection. These metrics highlight performance variability across different operational scenarios, especially under challenging conditions.

\textbf{Hyperparameter ablation.} To further evaluate the robustness of the model, the experiment was performed 10 times, with variations in the learning rate (LR), the momentum (M), the weight decay (WD), the number of convolutional (CONV) layers, and the number of batch normalization (BN) layers. The learning rate controls the size of the optimization step, while momentum integrates prior gradients for smoother updates. Weight decay was used for regularization to reduce overfitting. Changes in the number of convolutional and batch normalization layers affected feature extraction capacity and training stability. The hyperparameters for each run are detailed in Table~\ref{tab:hyperparams}. The results discussed earlier in this section are derived from run 2, which is the best performing run among the 10 runs. However, it is important to note that in runs 5 and 6, domain 6 shows a notable drop in accuracy. This may indicate that certain hyperparameter settings struggle with specific domain characteristics, highlighting the need for careful tuning to ensure robust performance across all domains. The combination of hyperparameters in run 2 allowed for fast convergence and minimal overfitting, balancing both learning rate and regularization. With a learning rate of 0.005, momentum of 0.85, and a relatively simpler model architecture with 2 convolutional layers and 1 batch normalization layer, this configuration optimized the learning process across all domains. With these hyperparameter settings, run 2 achieved the highest domain accuracy and F1-scores across most domains, demonstrating its effectiveness in domain-adaptive learning.

\begin{table}[h!]
    \centering
    \caption{Hyperparameters used across 10 runs. LR - Learning Rate, M - Momentum, WD - Weight Decay, CONV - Convolution Layers, BN - Batch Normalization Layers}
    \label{tab:hyperparams}
    \begin{tabular}{c c c c c c}
        \toprule
        \textbf{Run} & \textbf{LR} & \textbf{M} & \textbf{WD} & \textbf{CONV} & \textbf{BN} \\
        \midrule
        1  & 0.01  & 0.9  & 0.00001 & 3 & 2 \\
        \textbf{2}  & \textbf{0.005} & \textbf{0.85} & \textbf{0.00005} & \textbf{2} & \textbf{1} \\
        3  & 0.02  & 0.92 & 0.0001  & 4 & 3 \\
        4  & 0.015 & 0.88 & 0.00001 & 3 & 2 \\
        5  & 0.01  & 0.95 & 0.00002 & 5 & 4 \\
        6  & 0.008 & 0.87 & 0.00003 & 3 & 2 \\
        7  & 0.025 & 0.9  & 0.00005 & 4 & 3 \\
        8  & 0.012 & 0.93 & 0.00002 & 2 & 2 \\
        9  & 0.007 & 0.89 & 0.00001 & 4 & 3 \\
        10 & 0.001 & 0.91 & 0.00004 & 3 & 2 \\
        \bottomrule
    \end{tabular}%
\end{table}

\section{Conclusion}
\label{chap:Chapter5}

The primary objective of this work was to explore the application of online continual learning to predictive maintenance in railway systems, focusing on non-stationary environments. To achieve this, BOLT-RM (\textbf{B}oosting-inspired \textbf{O}nline \textbf{L}earning with \textbf{T}ransfer for \textbf{R}ailway \textbf{M}aintenance) was developed. The model’s ability to handle various train-track interactions, speeds and loads was evaluated through extensive experimentation, with results indicating that BOLT-RM significantly outperforms a isolated model approach. BOLT-RM achieved an average domain accuracy of 93\%, compared to 54\% for the isolated model, demonstrating its superior capacity to retain knowledge and generalize in domains. Forward transfer, which measures the model's ability to use past knowledge when learning new tasks, reached a value of 0.73, highlighting the strength of knowledge sharing between domains. The model also exhibited minimal forgetting, with a backward transfer score of nearly zero (\(3.47 \times 10^{-4}\)), confirming its ability to retain previously acquired knowledge without performance degradation. 

\textbf{Extensive ablations.} Experiments were conducted by varying the maximum number of domains used in the ER. The use of all 10 domains produced the highest average domain accuracy and the best forward transfer (around 95\%), probably due to the comprehensive usage of previous data. However, the 10-domain approach did not achieve the best backward transfer and also introduced significant computational overhead. In contrast, restricting replay to 5 domains achieved a balanced trade-off, with strong accuracy and notably low backward transfer (on the order of \(10^{-4}\)), confirming that the default 5-domain strategy remains an effective compromise between performance and efficiency. Alternative ER selection criteria were also studied, including mixing high- and low-loss domains (50\% highest loss and 50\% lowest loss, or 10\% highest and 90\% lowest, and vice versa). These approaches did not yield better results than focusing primarily on the highest-loss domains, which better support forward and backward transfer. Overall, these findings confirm that prioritizing the domains with the greatest difficulty (highest loss) provides the most robust strategy for handling non-stationary data in the PdM context. Following 10 experimental runs, the highest performance was achieved with a well-balanced configuration of learning rate and momentum, highlighting the importance of hyperparameter optimization. The precision, recall and F1-scores further emphasized the robustness of the model, particularly in domains 1, 2, and 7, where the F1-scores reached almost 1.00, indicating optimal fault detection. The statistical significance of the results was verified using the Friedman test (p-value of \(9.2 \times 10^{-4}\)). These findings underscore the effectiveness of the model in adapting to various operational conditions while also pinpointing areas for improvement in more complex domains.

\textbf{Research directions.} In future work, the main focus should be improving the dataset by integrating a wider variety of types, speeds, and operational scenarios to more accurately reflect real-world circumstances, as well as predicting more types of anomalies presented in the raw data. This broadening would facilitate more thorough evaluations of the flexibility and durability of the model in varied settings. Investigating these factors could help refine domain-specific strategies and improve sensor placement and data acquisition methods, ultimately enhancing the reliability and robustness of the model in the PdM of the railway wheel-track interface.

\bibliographystyle{cas-model2-names}

\bibliography{cas-refs}


\end{document}